\documentclass[final,5p,times,twocolumn]{elsarticle}

\journal{Journal of \LaTeX\ Templates}

\usepackage{numcompress}
\usepackage{graphics}
\usepackage{graphicx}
\usepackage{epsfig} 
\usepackage{amsmath} 
\usepackage{amssymb}  
\usepackage{bigstrut}
\usepackage{multirow}
\usepackage{booktabs}
\usepackage{threeparttable}
\usepackage{wrapfig,lipsum,booktabs}
\usepackage{color,soul}
\usepackage{setspace}
\usepackage{tabularx}

\usepackage[table]{xcolor}
\begin{document}

\begin{frontmatter}
\title{{Joint Prediction of Monocular Depth and Structure using Planar and Parallax Geometry}}

\author{Hao Xing\fnref{1}}
\author{Yifan Cao\fnref{1}}
\author{Maximilian Biber\fnref{2}}
\author{Mingchuan Zhou\fnref{3}\corref{*}}
\author{Darius Burschka\fnref{1}}
\fntext[1]{Authors are with Machine Vision and Perception Group, Technical University of Munich, Germany, email: hao.xing@tum.de, yifan.cao@tum.de, burschka@cs.tum.edu}
\fntext[2]{Author was with Zenuity GmbH and is with ML6, Berlin, Germany, email:  maximilianbiber@hotmail.com}
\fntext[3]{Author is with College of Biosystems Engineering and Food Science, Zhejiang  University, Hangzhou, China, email: mczhou@zju.edu.cn}
\cortext[*]{Corresponding author}


\newpage
\begin{abstract}

Supervised learning depth estimation methods can achieve good performance when trained on high-quality ground-truth, like LiDAR data. However, LiDAR can only generate sparse 3D maps which causes losing information. Obtaining high-quality ground-truth depth data per pixel is difficult to acquire. In order to overcome this limitation, we propose a novel approach combining structure information from a promising Plane and Parallax geometry pipeline with depth information into a U-Net supervised learning network, which results in quantitative and qualitative improvement compared to existing popular learning-based methods.
		
In particular, the model is evaluated on two large-scale and challenging datasets: KITTI Vision Benchmark and Cityscapes dataset and achieve the best performance in terms of relative error. Compared with pure depth supervision models, our model has impressive performance on depth prediction of thin objects and edges, and compared to structure prediction baseline, our model performs more robustly.

\end{abstract}


\end{frontmatter}


\graphicspath{{figures/}}

\section{Introduction}
\label{sec:intro}
    Depth estimation plays an important role in computer vision, which can be widely applied in robotics and autonomous driving. Monocular depth estimation is one of the most attractive tasks because of its low sensor cost, which predicts the depth value from a single RGB image. However, it is often described as an ill-posed task, since a single RGB image can be obtained by projection from an infinite number of 3D scenes \cite{LI2021108116}. 
	
	
    Supervised learning is one of the most widely applied solutions, which regard depth estimation as a regressive problem \cite{eigen2014depth,liu2015learning}. However, it is relying on highly precise pixel-wise ground-truth depth data, which is difficult to obtain, e.g., LiDAR sensor can only provide sparse high-precision depth measurement, and stereo-based vision sensors can only provide dense depth map for small detection range \cite{luo2019every, godard2019digging}.
    With sparse or low-precision ground-truth, it causes a limited quality of depth prediction. As one prominent example, after training with LiDAR data, the depth estimation of thin objects and edges is incorrect, due to missing label information of these areas in ground-truth data \cite{zhang2020depth}.
    
    The Plane and Parallax (P+P) geometry algorithm is an alternative solution, which can calculate the parallax of points that are not lying on the reference plane. With the given epipole of two images captured from different views, a ratio of height to the depth of a 3D point can be obtained, which can be converted to depth value for most pixels \cite{irani1996parallax}. However, the depth value around the epipole position is distorted, since the structure ratio tends to infinity, which is an inherent flaw of the P+P algorithm. {Depth values on dynamic objects (such as moving cars) are incorrectly predicted because the parallax of pixels on moving objects are contaminated by the relative motion \cite{wulff2017optical}.} Moreover, it fails when the ego-motion condition is not met, i.e., the epipole is at infinity \cite{sawhney19943d}. The supervised learning method can be a promising solution to solve the problems by optimizing the global images cost. 
	
    Hence, in this work, we propose a novel joint supervision network that applies U-Net architecture with two parallel output streams (structure and depth). The structure stream uses the output from the promising Plane and Parallax (P+P) geometry as the ground-truth, which provides the ratio between the height of each point above a planar surface and the depth in the camera scene. In a driving situation, the street plane can be approximated as a planar surface \cite{chaney2019learning}. The depth stream is using LiDAR measurement ground-truth. The structure ratio gives a strong and stable hint for depth regression and a combined cost of structure and depth forces the model to infer structure information around the epipole. When training the monocular depth model combined with structure information, the depth of thin objects and edges with sparse label information is optimally inferred to minimize the common loss function.
	
	Overall, the main contributions of our work lie in three fields:
	\begin{itemize}
		\item A novel structure and depth joint supervised network is proposed, which uses structure information as depth hints, and leverages depth and structure cost through a common loss function.
		\item We construct a promising 3D scene reconstruction pipeline using the Plane and Parallax algorithm to provide scene structure information from mono RGB images.
		\item Our model is evaluated on two large-scale and challenging dataset: KITTI Vision Benchmark \cite{geiger2012we} and Cityscapes dataset \cite{Cordts2016Cityscapes}. The result demonstrates that our model can not only enhance the monocular depth, but also structure estimation. We compare our results with other popular learning-based approaches on both datasets and achieve the best performance in terms of relative error, $0.054$ and $0.090$ respectively.
	\end{itemize}

\section{Related Work}

	Deep learning has evolved rapidly in recent years and redefined how the depth reconstruction problem is being approached. 
	The following section briefly introduces the most relevant methods, which differ from each other in their degree of supervision. Since this work builds upon the Plane+Parallax geometry, we also review the P+P framework.
	
	\subsection{Supervised Monocular Depth Estimation}
	Supervised methods directly regress disparity or depth from the input images. In doing so, ground-truth data is required to train a Convolutional Neural Network (CNN) as a regression model.
		
	Eigen \textit{et al.} \cite{eigen2014depth} are one of the first, who proposed a depth regression model that uses directly a single input image. Their work addresses integrating both global and local information of the scene to find depth relations from single images. In doing so, the authors present an architecture consisting of two neural networks that first generate initial depth predictions of the global scene, which are then locally refined at finer resolutions. Building upon this approach \cite{eigen2015predicting} and \cite{GeoNet} demonstrated the possibilities of using convolutional networks for multiple computer vision tasks. They extent the task of depth estimation by training a CNN to jointly predict different quantities, such as depth, surface norms, and semantic labels from the input image. This is done by either using a single CNN or using multiple CNNs that learn the relationship between the corresponding quantities while enforcing geometric consistency. In doing so, it was shown that it can be advantageous to combine highly related quantities that provide a rich representation of the scene and further improve each task’s estimations.
	
	Given the possibility of using a general regression model for various computer vision tasks, subsequent publication \cite{liu2015learning} focused on defining an appropriate training setup and loss function to further improve the accuracy and efficiency of supervised depth estimation from monocular images. 
	Fu \textit{et al.} \cite{fu2018deep} used a different approach and formulate the depth prediction task as an ordinal regression problem. In order to overcome the poor performance of depth supervision methods in edge detection, Feng \textit{et al.} \cite{XUE2021107901} firstly proposed a Boundary-inducted mechanism into a Scene-aggregated network, which improves the depth estimation performance of the area around boundary that indicates the depth change. 
	
    More recently, transformer networks have been successfully applied in supervised learning methods with incomplete ground truth \cite{ranftl2021vision}. Bhat \textit{et al.} \cite{bhat2021adabins} divided the depth range into bins through a transformer-based encoder and decoder network, which estimates depth value by the linear combination of bins center. Ye \textit{et al.} \cite{YE2021107578} proposed a two stream network that adopts a spatial attention module to extract pixel relationship combining with a depth regression branch, which significantly improved the performance of depth inferring.
	
	
	\subsection{Self-Supervised Monocular Depth Estimation}
	
	In comparison, self-supervised approaches only require image data for training. Garg \textit{et al.} \cite{garg2016unsupervised}, therefore, introduced a novel supervisory signal available from the geometric constraints between the input images themselves. The method uses the predicted depth of one image from a calibrated stereo pair to generate an inverse warp of the target image, which is then used to reconstruct the source image. The difference between the warped and the source image is treated as a training signal by computing the reconstruction loss. Subsequently, this approach was extended by reconstructing both input images, instead of only one \cite{godard2017unsupervised}. 
	
	Another type of self-supervised method generalizes earlier approaches to purely monocular settings. 
	Following established methods from Structure-from-Motion (SfM), \cite{zhou2017unsupervised} introduce a depth and pose network which is simultaneously learned from unlabeled monocular videos.
	Following \cite{garg2016unsupervised} the outputs are used to inversely warp the source images to reconstruct the target view. The photometric reconstruction loss is used to train the model.
	Since then, several methods \cite{luo2019every, wang2018learning} have been published building upon these approaches. To further improve this line of work, more advanced architectures or training processes were introduced \cite{godard2019digging, watson2021temporal}, which use additional loss terms, geometric constraints or generalizing to new datasets. Additionally, as already proven to work with supervised approaches, the depth estimation was extended to multiple computer vision tasks, such as joint depth, camera motion, optical flow, and motion segmentation learning. This set of improvements has shown that self-supervised approaches result in promising results by learning the depth from the geometric constraints instead of ground-truth data \cite{godard2019digging}. However, those methods suffer from scale ambiguity or occlusions, given that the supervisory signal does not contain absolute scale information \cite{watson2021temporal}.
	
	In most existing self-supervised learning methods, depth regression involves minimizing a photometric reconstruction loss. Finding the optimal depth value is typically difficult because the minimization of photometric reconstruction loss has multiple local minima, which leads to a decrease in prediction accuracy \cite{watson2021temporal}. 
	
    
	
	\subsection{Plane and Parallax geometry}
	
	The P+P geometry provides a powerful framework with a scene-centered representation of the 3D structure from 2D images. The main idea is to decompose the image motion between multiple frames into a planar homography and residual parallax, which can simplify the geometric reasoning about the scene structure \cite{irani1996parallax, sawhney19943d}. The underlying concept can be divided into two main steps. First, a planar surface is identified in the 3D scene and used to align the 2D images. This compensates for the detected planar motion. Second, the residual image displacement between the aligned frames is estimated. The displacements are either due to the parallax motion of static scene points or independently moving objects in the scene. This method provides dense structure information relative to an identified planar surface. 
	In many applications, the deviation relative to a planar surface in the scene is much more relevant than recovering the shape in terms of distance to the camera \cite{casser2019depth}. Although the P+P geometry provides a robust 3D reconstruction framework for most of the pixels, it can not predict a reasonable depth value at epipole because of the infinite structure ratio, and it predicts incorrectly depth value on dynamic objects due to parallax pollution from relative motion \cite{wulff2017optical}. Furthermore, it fails when there is no ego-motion \cite{sawhney19943d}.
	
	Inspired by the aforementioned works, we utilize additional dense structure information from P+P geometry to train a depth supervised model and try to achieve a better regression where there is no depth label. In doing so, we construct a 3D scene reconstruction pipeline using the Plane and Parallax algorithm to provide structure ground-truth, and feed both depth and structure streams to a joint supervised learning network and update the model through a common loss function.

\section{Joint prediction of structure and depth}
    
	Our depth and structure joint prediction pipeline consists of two separate parts. The first part is generating structure ground-truth data through our P+P implementation, see Fig.~\ref{fig:p+p}. It computes a dense projected structure map via the epipole from three consecutive input images. The reference plane is set as the ground surface. The second part contains the structure-guided training of monocular depth estimation using P+P geometry, see Fig~\ref{fig:joint}. It combines the depth and structure information by jointly learning both quantities through a fully shared decoder. Both predictions use the L1 loss and the total loss is computed as the weighted sum of depth and structure loss.

\subsection{Structure prediction using Plane and Parallax geometry approach}\label{sec:structure}

	\begin{figure*}[t!]
        \begin{centering}
        \begin{tabular}{cc}
            \includegraphics[width=0.98\textwidth]{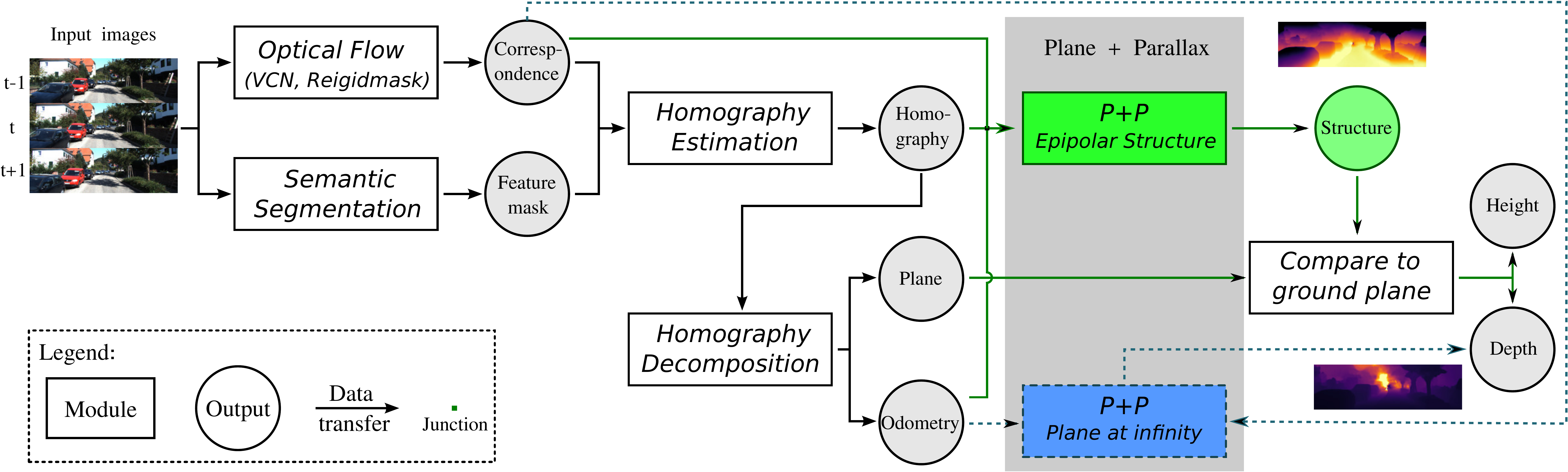}&
        \end{tabular}
        \caption{ \textbf{Overview of the Plane and Parallax pipeline:} Illustration of the structure prediction pipeline using Plane + Parallax geometry algorithm; The network compute the structure ground-truth data via the epipole from three consecutive input images with respect to the reference plane (solid green). It can predict directly depth, when set the reference plane at "infinity" (dash blue).}
        \label{fig:p+p}
        \end{centering}
    \end{figure*}

    From the theory of Plane and Parallax, it can be seen that a static scene and homography computing are essential to generate structure information. {The detail formula derivation} can be found in the Appendix~\ref{sec:appendix} or  \cite{irani1996parallax}.
    In doing so, we construct a P+P geometry based pipeline that includes \textit{optical flow}, \textit{semantic segmentation}, \textit{homography estimation}, \textit{homography decomposition} and \textit{P+P} modules, as shown in Fig.~\ref{fig:p+p}. We use an optical flow and semantic segmentation network to generate the point correspondences and the feature masks in the input frames. This information is required for computing temporarily and structurally consistent homographies, $^{t-1}A_{t}$ and $^{t+1}A_{t}$, to warp the adjacent frames $I_{t-1}$ and $I_{t+1}$ towards the current frame $I_{t}$ and to generate the structure value of each pixel using the derivations from \cite{irani1996parallax}.
	
	\textit{Optical Flow:} Since the warping of frames $I_{t-1}$ and $I_{t+1}$ by the homography towards current frame $I_{t}$ is highly dependent on accurate estimations of the point correspondences $p_{t-1}, p_{t},$ and $p_{t+1}$,
	we implement the optical flow network \textit{Rigidmask} \cite{yang2021learning} that achieves promising results on the KITTI Vision Benchmark for Optical Flow \cite{geiger2012we}. 
	The optical flow vectors are calculated in both directions (forwards and backwards): $^{t-1}u_{t}, ^{t+1}u_{t}, ^{t}u_{t-1},$ and $^{t}u_{t+1}$, with $^{t-1}u_{t} = p_{t-1}-p_{t}$. 
	
	\textit{Semantic Segmentation:} In the P+P framework, a reference surface needs to be defined to calculate the parallax. In this work, we choose the ground surface as the reference plane. In order to detect the ground surface, we apply the semantic segmentation network proposed in \cite{zhu2019improving}, which can distinguish one region from another based on its semantic context and is the leading method on the KITTI Vision Benchmark \cite{geiger2012we}. 
	
	\textit{Homography Estimation:} The optical flow vectors and the feature mask are subscribed by the homography estimation component to calculate the homography matrix $A$ that aligns the adjacent frames to the current frame. In this work, we implement \textit{MR-FLOW} method provided in \cite{wulff2017optical}, which employs an iterative scheme and alternates between optimization of structural components and the parallax based parameters. The relevant modules for computing the homography and optical flow refinement are used to achieve the given objective. The optical flow vectors are used to compute an initial homography estimation via RANSAC. The residual parallax vectors can be derived from the initial estimation by warping the point correspondences towards the reference frame $I_{t}$. Given the epipole, the homography is refined under the condition that the residual parallax vectors must intersect in the epipole. Therefore the distance of the residual parallax flow line to the epipole is minimized \cite{irani1996parallax}. 
	
	\textit{Homography Decomposition:} In addition to structural information, P+P can also use the odometry and plane information, obtained by decomposing the estimated homography, to directly predict the height relative to the reference plane and the depth of pixels. When the reference plane is set to be "at infinity", the depth is computed directly from the optical flow and odometry data (see dash blue branch in Fig.~\ref{fig:p+p}) without computing the structure as a result \cite{li2019learning}. We consider the depth prediction as a reference value to choose the best setup of the P+P network.
	
	However, the P+P algorithm is highly reliant on given epipole, static objects, and ego-motion. It fails around the epipole since the structure value tends to infinity, and predicts incorrect structure values on moving objects due to parallax pollution. As shown in Fig.~\ref{fig:p+p}, the homography is calculated using a static reference plane, and the homography result contains the information of ego motion. However, the dynamic objects in the scene have a relative motion with respect to the reference plane, which will have either shorter or longer parallax compared to static objects in that region and result in a wrong structure calculation using the same homography result as static objects. Furthermore, the P+P geometry fails when the camera has no ego-motion, since the homography is unsolvable. These errors will further lead to inaccurate depth prediction. More evidence can be found in quantitative and qualitative results.

\subsection{Joint prediction of depth and structure using Plane and Parallax}\label{sec:joint}

    \begin{figure*}[t!]
        \begin{centering}
        \begin{tabular}{cc}
            \includegraphics[width=0.80\textwidth]{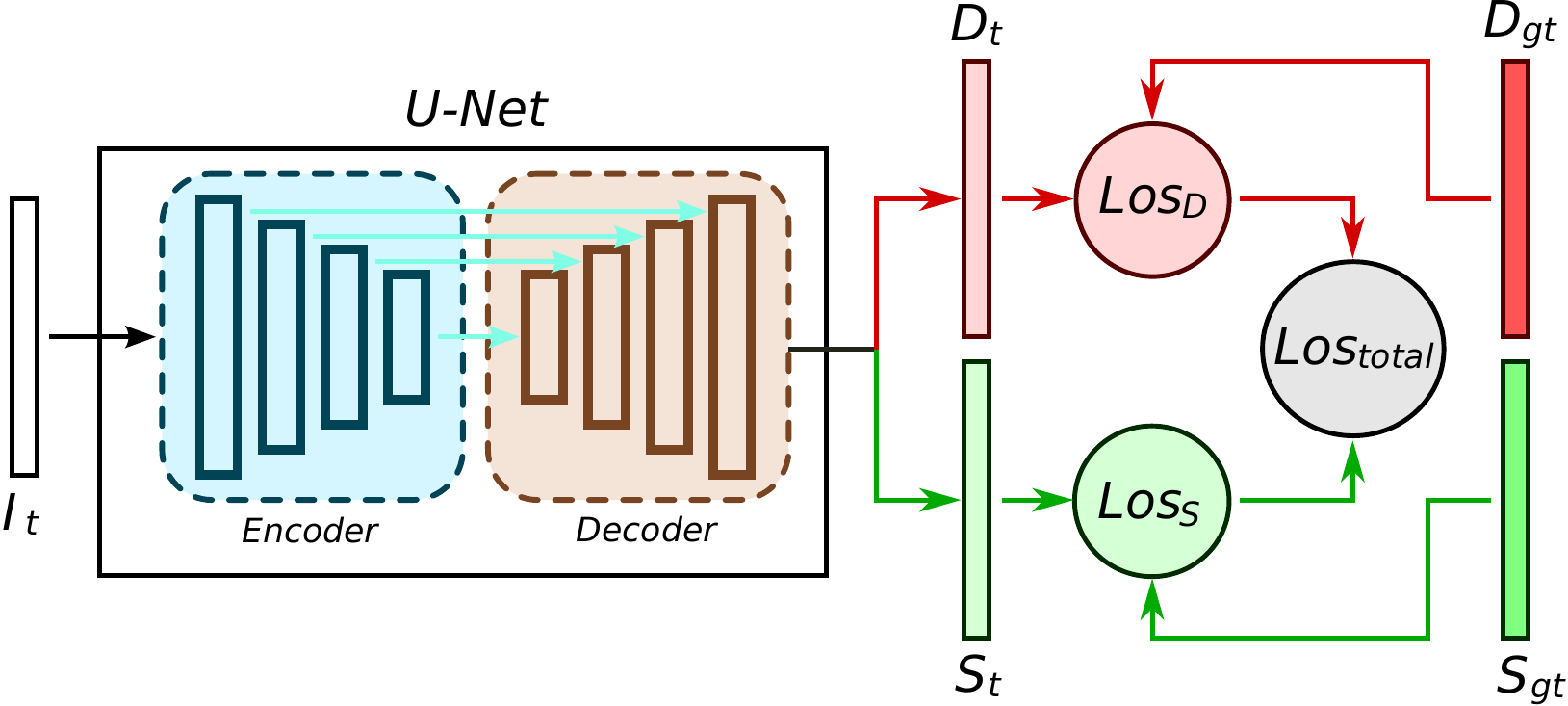}
        \end{tabular}
        \caption{ \textbf{Overview of the Depth+Structure joint supervision network:} Illustration of the joint prediction of depth and structure network, where $I_t$, $D_t$, $S_t$, $D_{gt}$, $S_{gt}$ stands for input, depth prediction, structure prediction, depth ground truth, structure ground truth, respectively.} 
        \label{fig:joint}
        \end{centering}
    \end{figure*}  
	
	To overcome the disadvantages of P+P geometry-based methods and pure depth supervised learning methods, we proposed a data-driven method that utilizes the output of the P+P framework. It combines the dense structure information provided by the P+P implementation with a supervised depth learning network to achieve two objectives: First, more precise prediction of dense depth where sparse LiDAR ground-truth fails to cover; Second, to gain robust structure information, even in the presence of moving objects or in static scenes, i.e., scenes without ego-motion.
	
	The architecture of the \textit{Depth+Structure} network builds upon a standard U-Net, as shown in Fig. \ref{fig:joint}. It uses the \textit{ResNeSt} \cite{zhang2020resnest} encoder at different complexities (50 and 101). The decoder consists of fully-shared weights among the depth and structure predictions and uses the sigmoid activation function at the output layer. In the depth branch, the sigmoid output $\sigma$, which corresponds to the disparity prediction of the network, is then converted to depth as it is done in \cite{godard2019digging} with $D = 1/(a\sigma +b)$ . In doing so, the depth output is constrained to lie in the range $\left[ 0.1,100\right]$, choosing a and b accordingly. In the structure stream, the range of the predicted structural values is defined by analyzing the full structure ground-truth that is generated by P+P geometry. It is found that $93.5\%$ of the structure images are of good quality and their absolute relative error is less than $0.5$. Within the range of these good quality data, the maximum structure value is found to be $0.06$ that is mainly located on the sides of the street, usually lower than middle street. The minimum value is found to be $-0.5$ that mostly appears in the top image area, e.g., the sky, tree tops, building tops and so on. Following this analysis,  the “mainly active” range of sigmoid output $\left[ \sigma(-2), \sigma(2) \right]$ is linearly mapped to the range $\left[-0.5,0.06\right]$. Note that the sign of structure value is defined by the normal vector of the reference plane (street). Since the normal vector is towards down, all scenes above the reference plane have negative values, otherwise positive.

	Both quantities are predicted with the same resolution, i.e., $640\times192$, and using the L1 loss. The total loss is computed by the weighted sum of two losses, as follows:
	\begin{equation}
		\begin{split}
		&\text{Loss} = w_{D}\text{Loss} _{D}+w_{S}\text{Loss} _{S} \\ &\text{with}
		\begin{cases}
			\text{Loss}_{D} = \sum_{i=1}^{n}\left\| D_{p}-D_{gt}\right\|\\
			\text{Loss}_{S} = \sum_{i=1}^{n}\left\| S_{p}-S_{gt}\right\|
		\end{cases}
		\end{split}
	\end{equation}
	where $D$ and $S$ represent the depth and the structure respectively, $p$ and $gt$ stand for predicted and ground-truth value, and $w$ is its corresponding weight. The ground-truth data of structure is obtained by the P+P implementation. The Depth+Structure network jointly predicts two quantities of different ranges to receive accurate predictions for both depth and structure. Therefore, an optimal ratio between the weighted structure and depth loss needs to be found, in the following denoted as $S/D$ ratio and calculated as $\frac{w_{S}}{w_{D}}$.

\section{Evaluation and Results}\label{Sec:4}
	To evaluate the performance of joint prediction of depth and structure, we experiment on two large-scale depth estimation datasets: KITTI dataset \cite{geiger2012we} and Cityscapes \cite{Cordts2016Cityscapes}. We first perform a detailed ablation study on the KITTI dataset to examine the contributions of the proposed model components to the depth and structure prediction performance. Then, we evaluate the final model on both datasets and compare the results with other state-of-the-art methods.

	\subsection{Dataset}
	
	\textbf{KITTI Dataset: }
	For the evaluation of the Plane+Parallax implementation, we use the KITTI Odometry dataset \cite{geiger2012we}. $2000$ monocular triplets of the provided sequences $00$-$10$ are randomly selected, representing approximately $10\%$ of each sequence, excluding sequence $01$. The sequence $01$ contains a highway scenario, for which the P+P method was not able to create meaningful results for the structure and depth predictions, due to inaccurate optical flow and homography predictions.
	
	For the experiment of the joint prediction network, we use the data split of Eigen \textit{et al.} \cite{eigen2015predicting}. The structure ground-truth data is provided by our implementation of P+P on KITTI Eigen split data. Given that the P+P implementation requires three consecutive images as input, the structure element cannot be created for all frames in the data split, 
	i.e., for images at the beginning and the end of a sequence. It is further necessary to remove static frames from the dataset, as ego-motion is required for the P+P formulation. We follow the pre-processing of Zhou \textit{et al.} \cite{zhou2017unsupervised} and generate $39,810$ monocular triplets. After implementation of the P+P method, we successfully create $38,798$ monocular triplets ground-truth data, in which $36,267$ with good quality ( REL $\leq0.5$ ) are used for training. 
	
	For our depth evaluation, we use the improved KITTI ground-truth data, which is created by an accumulation and data clean-up pipeline introduced by Uhrig \textit{et al.} \cite{uhrig2017sparsity}. This results in 652 semi-dense depth maps from the original 697 \textit{Eigen Test Split} data. In order to reduce the computation, we only use low-resolution ($640 \times 192$) images for the ablation study, then train the optimal model on high-resolution ($1024 \times 320$) images and compare the prediction results with existing top performers.
	
	\textbf{Cityscapes:} The Cityscapes dataset \cite{Cordts2016Cityscapes} is a more challenging dataset with many dynamic scenes. It is rarely used for depth prediction because of the lack of accurate ground truth. However, the dataset offers disparity images, which can be transferred to depth. With transferred depth ground truth, we used $17,598$ examples in training, $1,700$ images for evaluation, and $1,525$ samples for testing.


In this work, we report the absolute relative difference ($REL$), squared relative difference ($Sq\_Rel$), linear root mean squared error ($RMSE$), log root mean squared error ($RMSE_{log}$), threshold $\sigma$ and scale-invariant logarithmic error ($SILog$), which have been mostly employed in recent publications of monocular depth estimation \cite{godard2019digging, watson2019self, chaney2019learning}.

	
\subsection{Ablation study}
	
	\begin{table}[t]
		\caption{Evaluation of the Plane+Parallax implementation on 2.000 KITTI scenes*}
		\label{tab:ablation}

		\begin{center}
			\resizebox{1\linewidth}{!}{%
			\begin{tabular}{|l|l|l|c|c|}
				\hline
				OPTICAL FLOW	&  HOMOGRAPHY & METHOD & REL $\downarrow$ & REL$_{\text{w/o outliers}}$ $\downarrow$ \bigstrut \\ \hline\hline 
				\multirow{2}{*}{VCN KITTI}	& \multirow{2}{*}{OpenCV} & Structure  &0.1652  & 0.1302 \\ 
				\cline{3-5}
				& &Plane at infinity & 0.1453  & 0.1199 \\ \hline 
				\multirow{2}{*}{VCN General}	& \multirow{2}{*}{MR-FLOW} & Structure  &0.1698  & 0.1551 \\ 
				\cline{3-5}
				& &Plane at infinity& 0.1551 & 0.1257 \\ \hline 
                \multirow{2}{*}{VCN KITTI}	& \multirow{2}{*}{MR-FLOW} & Structure  & 0.1425  & 0.1424 \\ 
				\cline{3-5}
				& &Plane at infinity& 0.1296 & 0.1087 \\ \hline 
				\multirow{2}{*}{Rigidmask Mono}	& \multirow{2}{*}{OpenCV} & Structure  & 0.1394 & 0.1152 \\ 
				\cline{3-5}
				& &Plane at infinity& 0.1230 & 0.1038 \\ \hline 
				\multirow{2}{*}{Rigidmask Mono}	& \multirow{2}{*}{MR-FLOW} & Structure  &0.1085 & 0.1015 \\ 
				\cline{3-5}
				& &Plane at infinity&0.0987 & 0.0921 \\ \hline 
				\multirow{2}{*}{Rigidmask Stereo}	& \multirow{2}{*}{OpenCV} & Structure  &0.1365  & 0.1093 \\ 
				\cline{3-5}
				& &Plane at infinity& 0.1202 & 0.0990 \\ \hline 
				\multirow{2}{*}{Rigidmask Stereo}	& \multirow{2}{*}{MR-FLOW} & Structure  & 0.1077 & 0.0990 \\ 
				\cline{3-5}
				& &Plane at infinity& 0.0974 & 0.0974 \\ \hline 
				\multirow{2}{*}{Rigidmask Mono + Stereo}	& \multirow{2}{*}{MR-FLOW} & Structure  & \textbf{0.0936} & \textbf{0.0875} \\ 
				\cline{3-5}
				& &Plane at infinity&\textbf{0.0855} & \textbf{0.0794} \\ \hline 
			\end{tabular}
			}

		\scriptsize
		\begin{tablenotes}
		\item[a]$^*$ The evaluation metrics are based on the KITTI Odometry dataset. 
		\end{tablenotes}
		\end{center}
	\end{table}

	\begin{table*}[t]
	\vspace{0.12cm}
	\caption{Quantitative Results: Ablation study of the implemented joint depth and structure predictions.}
	\label{tab:s/d}
	\begin{center}
		\resizebox{\linewidth}{!}{%
			\begin{tabular}{|c|c|c|c|c|c|c|c|c|c|c|c|c|}
				\hline
				\multirow{2}{*}{\textbf{S/D ratio}}	&  \multicolumn{2}{c|}{\textbf{Encoder}} & \multicolumn{2}{c|}{\textbf{Decoder}$^a$} & \multicolumn{8}{c|}{\textbf{Evaluation Metrics -  KITTI (improved Eigen Test Split)}$^b$} \\
				\cline{2-13}
				&50 &101 & f-s & n-s & REL $\downarrow$ &Sq Rel $\downarrow$& RMSE $\downarrow$ & $\text{RMSE}_{\text{log}} \downarrow$ & $\delta< 1.25 \uparrow$ & $\delta< 1.25^2 \uparrow$ & $\delta< 1.25^3 \uparrow$ & SILog $\downarrow$ \\ \hline\hline
			
				\multirow{4}{*}{1}& $\times$ & & $\times$& &0.0649 & \underline{0.2929} & \underline{2.9144} &0.1049 & \underline{0.9413} & \underline{0.9899} & 0.9975& \underline{9.6833} \\ \cline{2-5}
				& $\times$ & & & $\times$ & \underline{0.0647} & {0.2976} & 2.9533 & \underline{0.1047} & 0.9432 & 0.9896 & \underline{0.9976} & 9.6723 \\ \cline{2-13}
				&  & $\times$ & $\times$ & & 0.0615 & 0.2811 & \underline{2.8644} & \underline{0.0999} & 0.9484 & \underline{0.9910} & \underline{0.9978} & 9.1863 \\ \cline{2-5}
				&  &$\times$  & &$\times$ & \underline{0.0614} & \underline{0.2798} & 2.8646 & 0.1000 & \underline{0.9487} & 0.9909 & \underline{0.9978} & \underline{9.1709} \\
				\specialrule{.1em}{.05em}{.05em}
				\multirow{4}{*}{10}& $\times$ & & $\times$ & & 0.0636 & 0.2853 & 2.8766 & 0.1023 & 0.9446 & 0.9909 & \underline{0.9979} & 9.4479 	\\ \cline{2-5}
				& $\times$ & & & $\times $ & 0.0645 & 0.3009 & 2.9801 & 0.1048 & 0.9418 & 0.9898 & 0.9975 & 9.6666	\\ \cline{2-13}
				&  & $\times$ & $\times$& & \underline{\textbf{0.0604}} & \underline{0.2774} & 2.8719 & \underline{\textbf{0.0991}} & \underline{\textbf{0.9501}} & 0.9908 &0.9978 & \underline{\textbf{9.1116}}	\\ \cline{2-5}
				&  &$\times$ & & $\times$ &0.0611 & 0.2791 & \underline{2.8653} & 0.0996 & 0.9486 & \underline{\textbf{0.9911}} & \underline{0.9979} & 9.1634	\\ 
				\specialrule{.1em}{.05em}{.05em}
				\multirow{4}{*}{100}& $\times$ & & $\times$ & & 0.0643 & 0.2848 & 2.9109 & 0.1037 & 0.9450 & 0.9898 & 0.9978 & 9.6500  \\ \cline{2-5}
				& $\times$ &  &  &$\times$ & 0.0633 & 0.2826 & 2.9179 & 0.1027 & 0.9459 & 0.9906 & 0.9979 & 9.5028  \\ \cline{2-13}
				&  & $\times$ &  $\times$& & \underline{0.0613} & 0.2760 & 2.8893 & 0.0998 & 0.9495 & 0.9908 & 0.9979 & 9.2152 \\ \cline{2-5}
				&  &$\times$  &  &$\times$ & 0.0614 & \underline{\textbf{0.2747}} & \underline{\textbf{2.8640}} & \underline{{0.0993}} & \underline{\textbf{0.9501}} & \underline{0.9909} & \underline{\textbf{0.9981}} & \underline{9.1599}  \\ 
				\hline
			\end{tabular}
			}
		\end{center}
		\scriptsize
		\begin{tablenotes}
    		\item[a]$^a$ The decoder configurations are denoted as: "f-s" = fully-shared; "n-s" = non-shared.
    		\item[b]$^b$ The evaluation metrics are based on the \textbf{KITTI Eigen Zhou test dataset} with resolution of $640\times192$. The best results comparing all \textit{S/D} ratios are in \textbf{bold}; The best results of the encoder-decoder setup for each \textit{S/D} ratio are \underline{underlined}.
		\end{tablenotes}
	\end{table*}

    \textbf{Plane+Parallax implementation:} In order to obtain the best structure ground-truth data, we execute the experiment of P+P with different optical flow networks and homography estimation tools. Since there is no benchmark for the projective structure, we convert the computed structure to depth using the plane parameters, see Fig. \ref{fig:p+p}. The predictions are compared against the LiDAR ground-truth using the mean absolute relative error. The results in Table \ref{tab:ablation} confirm that the optical flow accuracy is of high significance for the resulting depth maps. 
    According to the KITTI Vision Benchmark for Optical Flow \cite{geiger2012we}, the optical flow network \textit{Rigidmask} scores better than the alternative network \textit{VCN}. This can also be seen in Table \ref{tab:ablation}. 
    
	
	By comparing the metrics produced by the two homography estimation tools, \textit{OpenCV} and \textit{MR-FLOW}, it is clear that the \textit{MR-FLOW} facilitates significantly more accurate estimates, which includes the parallax constraints in the homography estimation to enforce consistent predictions over multiple frames.

    Another remarkable observation is that the approach setting the planar surface to infinity, which is denoted as \textit{Plane at infinity}, consistently outperforms the depth predictions resulting from the structure, even though both approaches use the same odometry data. 
    This indicates that the position of the reference plane affects the quality of the resulting depth maps. 

	\begin{table}[t]
		\centering
		\caption{Performance analysis of the U-Net on the KITTI dataset with varying backbone encoder$^*$}
		\label{tab:encoder}
		\begin{center}
			\resizebox{1\linewidth}{!}{%
			\begin{tabular}{|l|c|c|c|c|c|c|}
				\hline
				Encoder	& REL $\downarrow$ & RMSE $\downarrow$ & $\text{RMSE}_{log} \downarrow$  &
				$\delta^1 \uparrow$ & $\delta^2 \uparrow$ & $\delta^3 \uparrow$ 
				\bigstrut \\ \hline\hline 
			    ResNet101 \cite{he2016deep} & 0.078 & 3.457 & 0.120 & 0.925 & 0.984 & 0.996 \\ \hline
				ResNext101 \cite{xie2017aggregated}& 0.070 & 3.217 & 0.112 & 0.938 & 0.988 & 0.997 \\ \hline
				ResNeSt101 \cite{zhang2020resnest}& \textbf{0.068} & \textbf{3.083} & \textbf{0.108} & \textbf{0.943} & \textbf{0.990} & \textbf{0.998} \\ \hline
			\end{tabular}}
		    \scriptsize
		    \begin{tablenotes}
		    \item[a]$^*$ The evaluation metrics are based on the KITTI Eigen Zhou test dataset with improved ground-truth. 
		    \end{tablenotes}
		\end{center}

	\end{table}

Based on the experiments, the configuration using the optical flow \textit{Rigidmask Stereo}, combined with the consistent homography estimation via \textit{MR-FLOW}, can be seen as the most promising one.

\textbf{Pure depth supervised learning method:} To confirm the effectiveness of the proposed model (Depth+Structure), we also conduct various benchmark experiments on the depth estimation baseline model by changing backbone encoders with three frameworks,i.e., \textit{ResNet}$101$, \textit{ResNext}$101$ and \textit{ResNeSt}$101$. The corresponding results are shown in Table~\ref{tab:encoder}, in which ResNeSt achieves the best performance.

\textbf{Jointly estimation of depth and structure:} To analyze the influence of the additional structure input on the depth prediction and to evaluate the performance of the structure prediction itself, we execute the experiments concerning different configurations of the \textit{S/D} ratio, fully shared and non-shared decoder, and complexity of the encoder. Table \ref{tab:s/d} summarizes the evaluation results, including various encoder-decoder configurations.

The \textit{S/D} ratio is a trade-off parameter. After $\textit{S/D}=10$, the more weight is applied to the structure loss, the worse the result of absolute relative error. 
This observation leads to the conclusion that the structure ground-truth contributes to a better depth estimation with a suitable \textit{S/D} loss weight ratio. Both less and much structure information may hamper the training process, which precludes the network from precisely predicting two separate quantities simultaneously. Comparing all depth metrics it can be seen that the model with an \textit{S/D} ratio of $10$ results in the best performance of absolute relative error. The model with an \textit{S/D} ratio of $1$, performs the worst overall metrics.

The experimental results concerning the loss weighting demonstrates that the model might not be able to jointly predict the depth and structure of the same accuracy. This motivates to increase the complexity of both encoder and decoder and investigate whether this leads to further improvements. 

\begin{table}[t]
    \centering
    \caption{The number of parameters of different encoder-decoder setups.}
    \label{tab:num_param}
    \begin{center}
    \resizebox{\linewidth}{!}{%
			\begin{tabular}{|c|c|c|c|c|}
			\hline
			Encoder & Decoder & Encoder Parameters & Decoder Parameters & Total Parameters \\ \hline\hline
			\multirow{2}{*}{\textit{ResNeSt50}} & \textit{Fully-shared} & $27.483.240$ & $9.012.081$ & $36.495.321$ \\ \cline{2-5}
			& \textit{Non-shared} & $27.483.240$ & $18.024.162$ & $45.507.402$ \\ \hline
			\multirow{2}{*}{\textit{ResNeSt101}} & \textit{Fully-shared} & $48.275.016$ & $9.012.081$ & $57.305.674$ \\ \cline{2-5}
			& \textit{Non-shared} & $48.275.016$ & $18.024.162$ & $66.336.332$ \\ \hline
			\end{tabular}
			}
    \end{center}
\end{table}

\begin{table*}[t]
	\caption{\textbf{Class-specific evaluation:} Comparison of the Depth Baseline (D) and the
Depth+Structure (D+S) model*.}
	\label{tab:4}
	\begin{center}
		\resizebox{\linewidth}{!}{%
			\begin{tabular}{|c|c|c|c|c|c|c|c|c|c|}
				\hline
				Class	&  Method & REL $\downarrow$ & Sq Rel $\downarrow$ &RMSE $\downarrow$ & $\text{RMSE}_{\text{log}}\downarrow$ & $\delta< 1.25\uparrow$ & $\delta< 1.25^2\uparrow$ & $\delta< 1.25^3\uparrow$ & SILog $\downarrow$ \\ \hline\hline
				\multirow{2}{*}{Street}	& D & 0.0389 & 0.0769 & 1.2898 & 0.0554 & 0.9899 & 0.9985 & 0.9997 & 4.1763 \\
				& D+S & \textbf{0.0321} & \textbf{0.0555} & \textbf{1.0906} & \textbf{0.0456} & \textbf{0.9930} & \textbf{0.9993} & \textbf{0.9999} & \textbf{3.6099} \\ \hline
				\multirow{2}{*}{Fence}	& D & 0.1682  &1.5062  & 5.1325 & 0.1962 & 0.7608 & 0.9316& 0.9802& 11.7992 \\
				& D+S & \textbf{0.1167} & \textbf{0.9696} & \textbf{4.2157} & \textbf{0.1453} & \textbf{0.8455} & \textbf{0.9539} & \textbf{0.9916} & \textbf{9.4513} \\ \hline
				\multirow{2}{*}{Thin Objects}	& D & 0.2390  &2.6241  & 7.6115 & 0.2729 & 0.6619 & 0.8806& 0.9509& 21.4924 \\
				& D+S & \textbf{0.1391} & \textbf{1.4043} & \textbf{5.8517} & \textbf{0.1738} & \textbf{0.8095} & \textbf{0.9485} & \textbf{0.9874} & \textbf{13.5015} \\ \hline
				\multirow{2}{*}{ Walking Person + Rider}	& D & 0.1868 & 1.7838 & \textbf{5.4203} & 0.2062 & 0.7347 & 0.9212 & 0.9788 & 12.3768 \\
				& D+S & \textbf{0.1451} & \textbf{1.4734} & {5.5421} & \textbf{0.1728} & \textbf{0.7988} & \textbf{0.9482} & \textbf{0.9908} & \textbf{10.6492} \\ \hline
				\multirow{2}{*}{Car (Static)}	& D & 0.1125 & 0.9713 & 4.7381 & 0.1674 & 0.8822 & 0.9660 & 0.9873 & 14.3663 \\ 
				& D+S & \textbf{0.0901} & \textbf{0.7342} & \textbf{3.9863} & \textbf{0.1295} & \textbf{0.8908} & \textbf{0.9775}& \textbf{0.9907} & \textbf{9.7522} \\ \hline
				\multirow{2}{*}{Car (Dynamic)}	& D & 0.1143 & \textbf{1.0213} & \textbf{4.8286} & 0.1677 & 0.8742 & 0.9627 & 0.9849 & 13.3413 \\ 
				& D+S & \textbf{0.0983} & {1.0357} & {5.2197} & \textbf{0.1283} & \textbf{0.8988} & \textbf{0.9781}& \textbf{0.9907} & \textbf{9.4993} \\ \hline 
				\multirow{2}{*}{Edges of Car (Static)} & D & 0.1367 & 1.2024 & 6.0378 & 0.1972 & {0.8322} & 0.9524 & 0.9828 & 17.5617 \\
				& D+S & \textbf{0.0995}  & \textbf{0.8651} & \textbf{4.6086} & \textbf{0.1450} & \textbf{0.8742} & \textbf{0.9719}& \textbf{0.9887} & \textbf{11.2698} \\ \hline
				\multirow{2}{*}{Edges of Car (Dynamic)}	& D & {0.1488} & 1.7931 & {7.3208} & 0.2117 & {0.7999} & 0.9355 & 0.9786 & {17.4073} \\
				& D+S & \textbf{0.1040} & \textbf{1.1729} & \textbf{5.6866} & \textbf{0.1378} & \textbf{0.8902} & \textbf{0.9737}& \textbf{0.9889} & \textbf{10.6479} \\ \hline
				
		\end{tabular}}
	\end{center}
	\scriptsize
	\begin{tablenotes}
        \item[a]$^*$ The evaluation metrics are based on the \textbf{KITTI Eigen Zhou test} with improved ground-truth. The best results of each class are in \textbf{bold}. D+S is the model using ResNeSt 101 encoder, full-shared decoder and S/D = 10. The resolution for training is $640 \times 192$.
        
    \end{tablenotes}
\end{table*}

The effect of increasing the encoder’s complexity can be observed by comparing the results in Table \ref{tab:s/d}. Using a more complex encoder (\textit{ResNeSt101}), facilitates better performances across all metrics and weighting configurations. 
The comparison of the two decoder setups, i.e., \textit{fully-shared} weights compared to \textit{non-shared} weights, is of particular interest. As long as one of the losses dominates the overall loss, such as with \textit{S/D} ratio of $1$ and $100$, the non-shared decoder performs slightly better than the fully-shared decoder. When the \textit{S/D} ratio is $10$, it achieves the best performance. These observations motivate us to further analyze this setup in the qualitative analysis. The number of parameters of different encoder-decoder setups can be found in Table~\ref{tab:num_param}, in which the \textit{fully-shared} decoder has the half size of \textit{non-shared} decoder. Our best model has 57.3M parameters, in which the \textit{ResNeSt101} encoder has 48.3M parameters, and the \textit{fully-shared} decoder has 9.0M parameters.



The class-specific evaluation reveals some of the potential benefits of including the structure predictions in the training process. Due to the sparse ground-truth data, the depth baseline usually cannot predict classes that contain fine-grained structures and sharp boundaries in the scene, such as poles, traffic signs, and traffic lights. As evident from Table \ref{tab:4}, the Depth+Structure model performs better across all metrics. Additionally, the results of the class ``Person", ``Car" and ``Edges of Car" show that the considered structure information leads to a better representation of the scene and, therefore, to a safer implementation in the field of autonomous driving. The comparisons between the  ``Static" and  ``Dynamic" scenes certify a drawback of monocular depth prediction, which cannot predict well the depth value of a fast-moving object using blurred RGB images. 
The performance degradation of our method switching from ``Car (Static)" to ``Car (Dynamic)" reflects the impact of the natural imperfection of P+P geometry on joint learning, which cannot correctly predict structure values of moving objects. On the other hand, it also confirms that our joint learning-based model can compensate for the drawback of P+P geometry and successfully predict reasonable structure values of dynamic objects.
The class-specific evaluation further exemplifies the significance of fine-grained metrics for depth evaluation.
	
	\begin{figure}[t]
		\centering
		\includegraphics[width=0.95\linewidth]{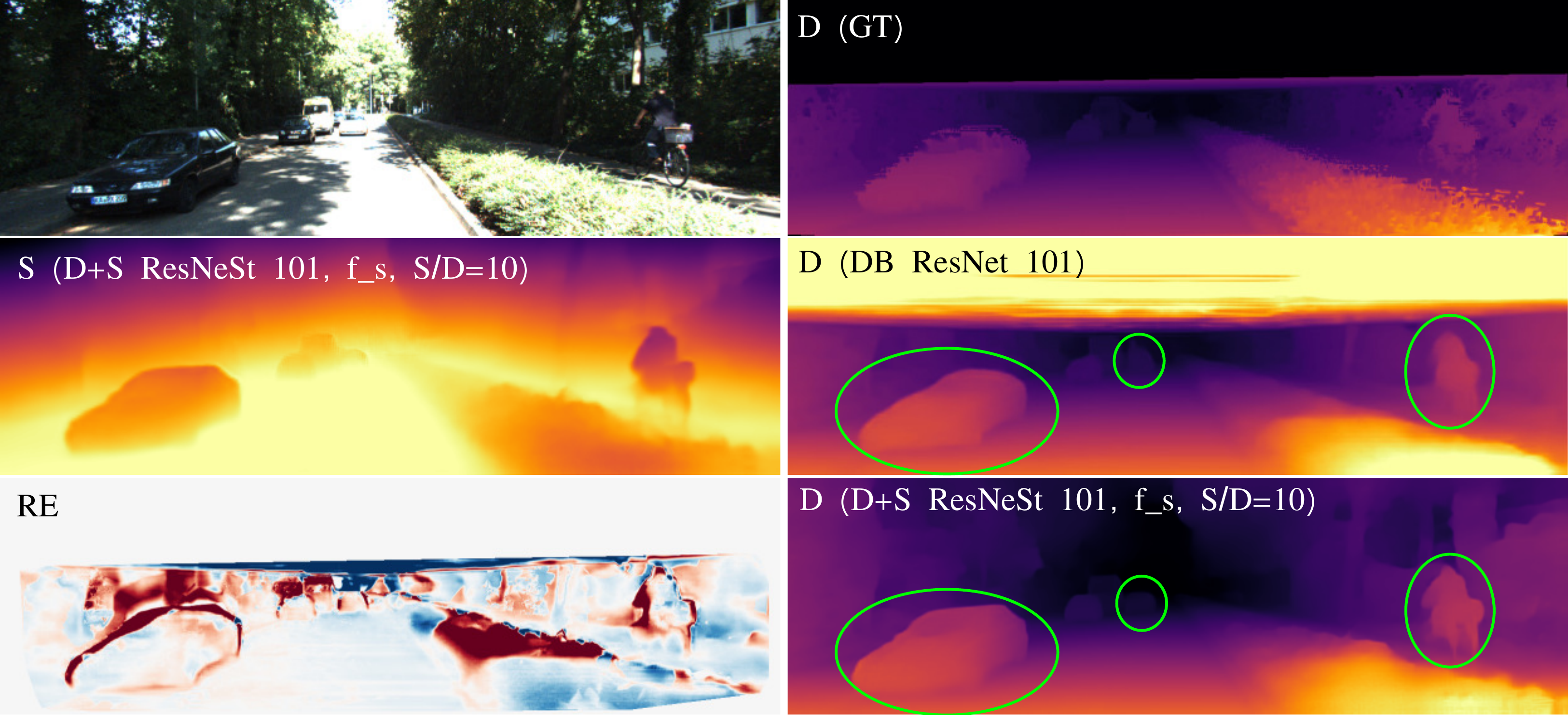}
		\caption{\textbf{Qualitative Results:} Row 1: Original image, and the depth (D) ground-truth (GT). Row 2: Structure prediction (S) using our joint model (D+S), and depth prediction of depth baseline (DB). Row 3: Difference of the relative errors in depth (RE) and the depth prediction of the joint model (D+S). Red region: Depth+Structure model performs better; Blue region: Depth Baseline performs better.}
		\label{fig:qualitative}
	\end{figure}
	
	\begin{figure}[t]
		\centering
		\includegraphics[width=0.95\linewidth]{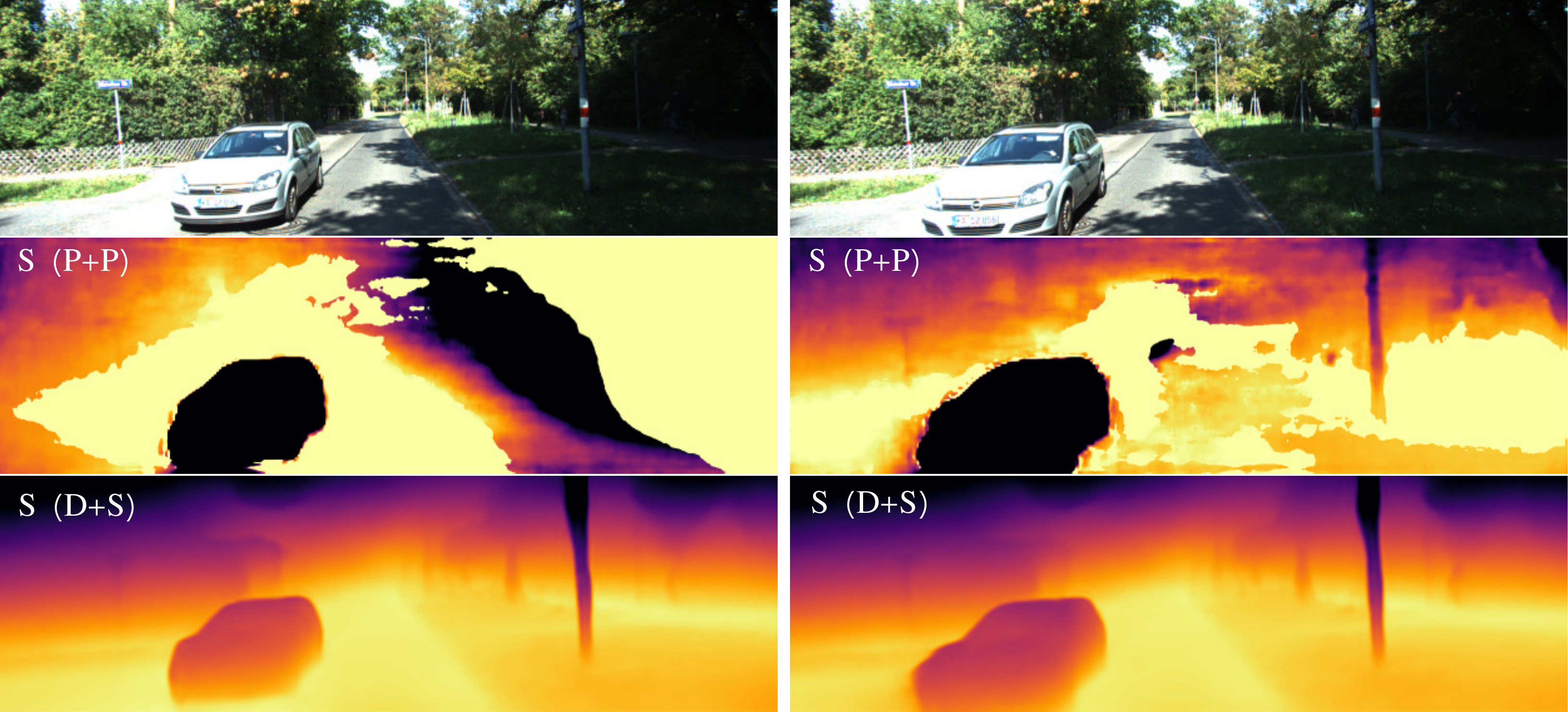}
		\caption{\textbf{Enhanced structure prediction by the joint model under extreme condition:} Row 1: Original images. Row 2: Structure prediction of P+P. Row 3: Structure prediction of our joint model D+S.}
		\label{fig:qualitative2}
	\end{figure}
	
	\textbf{Qualitative  Results:} The scene in Fig. \ref{fig:qualitative} contains one oncoming car in front of the ego-vehicle, one rider on the right side, several stop cars on the left side of the image, and multiple thin objects, like trees and poles. Comparing the baseline depth prediction confirms the results of the class-specific evaluation, namely that the Depth+Structure model performs better for the  ``Thin Objects". Whereas the baseline model misses the oncoming car in the front, wrongly predicts the edge of the stopped car and the rider, those are clearly visible in the jointly trained model’s depth predictions. 
	Here the influence of the structure predictions during training becomes evident. The structure prediction contains relevant information about the objects in the scene, which are shared with depth prediction through the fully-shared decoder.
	
	The difference of the relative errors of each depth prediction supports these results: the positive values (red) show the areas in which the joint depth and structure predictions yield better results; for the negative values (blue) the baseline model performs better. 
	\begin{figure}[t!]
		\centering
		\includegraphics[width=0.95\linewidth]{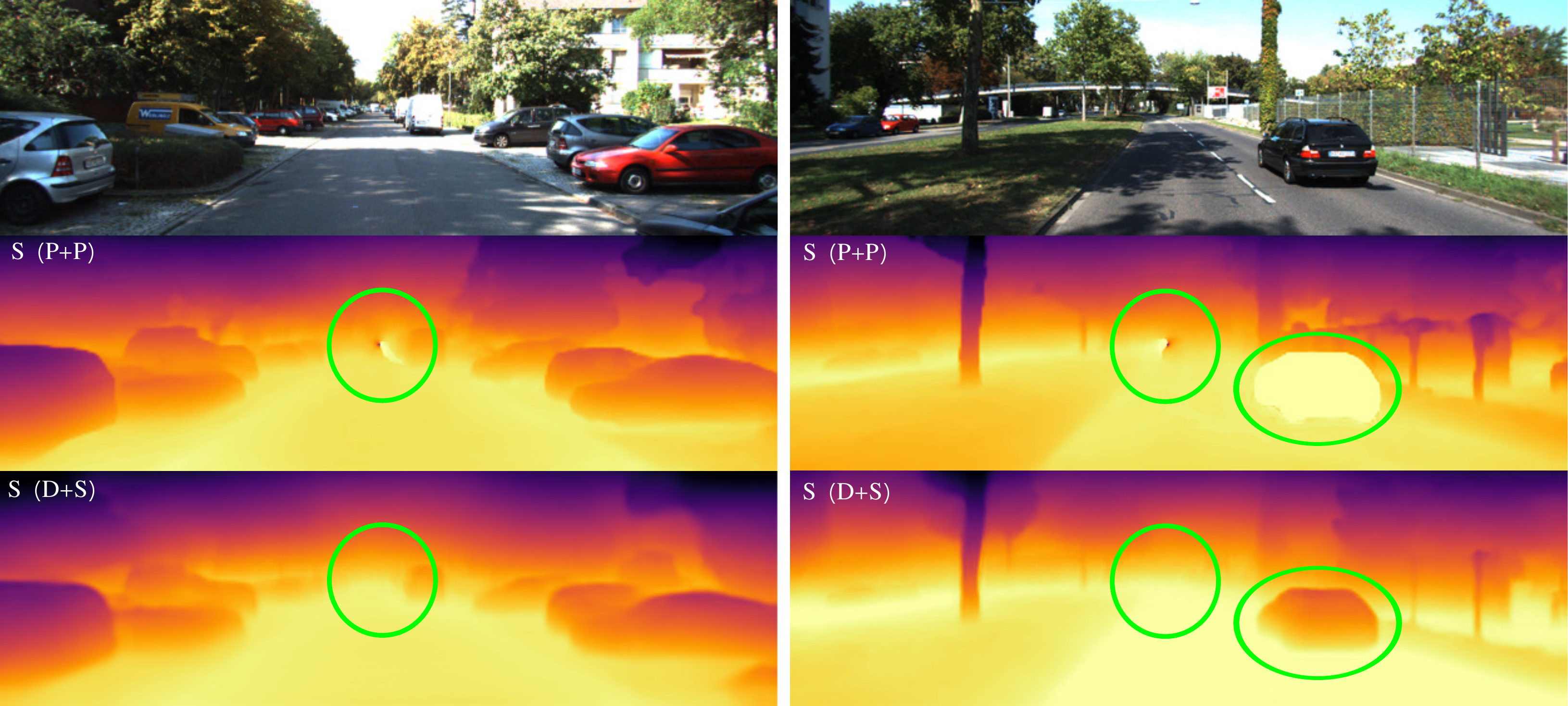}
		\caption{\textbf{Enhanced structure prediction by the joint model around epipole area (circle):} Row 1: Original images. Row 2: Structure prediction of P+P. Row 3: Structure prediction of our joint model D+S.}
		\label{fig:qualitative3}
	\end{figure}

\begin{table*}[t!]
	\centering
	\caption{Quantitative results on KITTI dataset: Comparison with top performer$^a$}
	\label{tab:state_of_the_art}
	\begin{center}
		\resizebox{\linewidth}{!}{%
			\begin{tabular}{|c|c|c|c|c|c|c|c|c|c|}
				\hline
				Method$^b$	&  Train$^c$ & \#Params &REL $\downarrow$ & Sq Rel $\downarrow$ &RMSE $\downarrow$ & $\text{RMSE}_{\text{log}} \downarrow$ & $\delta_1 \uparrow$ &  $\delta_2 \uparrow$ & $ \delta_3 \uparrow$ \\ \hline\hline
				
				SfMLearner \cite{zhou2017unsupervised}	& M & -&0.176  &1.532  & 6.129 & 0.244 & 0.758 & 0.921& 0.971 \\ 
				GeoNet \cite{yin2018geonet} & M & -&0.132& 0.994& 5.240& 0.193&0.833& 0.953& 0.985 \\
				DDVO \cite{wang2018learning}&M & -&0.126& 0.866 &4.932& 0.185& 0.851 &0.958 &0.986\\
				Ranjan \cite{ranjan2019competitive} & M & -&0.123& 0.881& 4.834 &0.181& 0.860 &0.959& 0.985\\
				EPC++ \cite{luo2019every} & M & -& 0.120& 0.789& 4.755 &0.177& 0.856& 0.961 &0.987 \\
				Monodepth2 \cite{godard2019digging} &M& -& 0.090 &0.545 &3.942& 0.137& 0.914& 0.983& 0.995 \\
				ManyDepth \cite{watson2021temporal} & MS & -& {0.058}  & {0.334}  & {3.137} & {0.101} & {0.958} & {0.991}& {0.997}\\ 
				ManyDepth ($ 1024 \times 320 $) \cite{watson2021temporal} & MS & -& {0.055}  & {0.305}  & {2.945} & {0.094} & {0.963} & {0.992}& {0.997}\\
				\hline
				
				Eigen \textit{et al.} \cite{eigen2014depth}	& D & 83M & 0.190  &1.515  & 7.156 & 0.270 & 0.692 & 0.899& 0.967 \\ 
				Liu \textit{et al.} \cite{liu2015learning} & D & 20M & 0.217  & 1.841 & 6.986 & 0.289 & 0.647 & 0.882& 0.961  \\ 
				Baseline ResNet101 & D & 54M & 0.078  & 0.416& 3.457 & 0.126 & 0.925 & 0.984& {0.996} \\ 
				
				DORN ($1024 \times 320$, pretrained)~\cite{fu2018deep} & D & 100M & {0.072}  &{0.307}  & {2.727} & {0.120} & {0.932} & 0.984& 0.994\\ 
				Baseline ResNeSt101 & D & 57M & 0.068  & 0.309& 3.083 & 0.108 & 0.943 & 0.990 & {0.998} \\ 
				DPT-Hybrid ($ 1024 \times 320$, pretrained)~\cite{ranftl2021vision} & D & 123M & {0.062}  & -  & {2.573} & {0.092} & {0.959} & \textbf{0.995} & \textbf{0.999}\\
				AdaBins ($ 1024 \times 320$, pretrained)~\cite{bhat2021adabins} & D & 78M & {0.058}  & -  & \textbf{2.360} & \textbf{0.088} & \textbf{0.964} & \textbf{0.995} & \textbf{0.999}\\
				\hline


				\textbf{Ours}$^d$ (\textit{VCN KITTI})& D+S & 57M & {0.078} & 0.399 & 3.373 & {0.125} & {0.926} & {0.985} &{0.996} \\
				\textbf{Ours}$^d$ (\textit{ResNet101})& D+S & 54M &  {0.065} & 0.321 & 3.099 & {0.106} & {0.940} & {0.985} &{0.997} \\
				\textbf{Ours}$^d$ ($640 \times 192$) & D+S & 57M &  {0.060} & 0.277 & 2.872 & {0.099} & {0.950} & {0.991} &{0.998} \\  
				\textbf{Ours}$^d$ ($1024 \times 320$)& D+S & 57M &  {0.056} & 0.230 & 2.617 & {0.091} & {0.957} & {0.993} &{0.998} \\ 
				\textbf{Ours}$^d$ ($1024 \times 320$, pretrained)& D+S & 57M & \textbf{0.054} & \textbf{0.210} & {2.467} & \textbf{0.088} & {0.961} & \textbf{0.995} &\textbf{0.999} \\ 
				\hline
		\end{tabular}}
	\end{center}
	\scriptsize
	\begin{tablenotes}
		\item[a]$^a$ The evaluation metrics are based on the \textbf{KITTI Eigen Zhou test} with improved ground truth. The best results of each class are in \textbf{bold}. 
		\item[b]$^b$ The default resolution for training is $640 \times 192$. \emph{DORN} uses pretrained \emph{VGG-16} and \emph{ResNet101} as encoder. \emph{DPT-Hybrid} is pretrained on three datasets, \emph{AdaBins} uses pretrained \emph{EfficientNet B5} encoder.
		\item[c]$^c$ D refers to methods that use depth supervision at training time. M is for models trained on mono data. S means using stereo images. D+S refers to the methods trained on both mono depth and structure information. 
		\item[c]$^d$ Our model is using structure ground-truth of Rigidmask Mono + Stereo P+P geometry, a ResNeSt-101 encoder, and a full-shared decoder with $\text{S/D} = 10$, if not stated in the remarks.
	\end{tablenotes}
\end{table*}

	Fig. \ref{fig:qualitative2} and Fig. \ref{fig:qualitative3} give a comparison of structure prediction between the P+P geometry framework and our joint learning-based model. In Fig. \ref{fig:qualitative2}, the selected scenes show two successive images in which the ego-vehicle is static while waiting for the approaching car to pass. The example reveals two limitations of the Plane+Parallax implementation that are eliminated by the data-driven approach: first, the moving car cannot be predicted in the Plane+Parallax implementation since the independent motion violates the parallax constraints; second, the system also fails to predict the static background in this scene due to the absence of ego-motion. From the last row, it can be seen that these limitations are overcome by our \textit{Depth+Structure} model. 
	In Fig. \ref{fig:qualitative3}, we compare the performance of the P+P geometry with our joint learning-based method on static scenario (left side) and dynamic scenario (right side), where the static scenario contains only static objects and the dynamic scenario contains a car driving right ahead. In both scenarios, the P+P geometry framework fails around the epipole, while the joint model successfully infers the structure value at epipole. In the dynamic scenario, the structure values of dynamic object (moving car) are incorrectly predicted by P+P geometry and are the smallest in the whole scene, since the parallax are shortened by the relative motion, while our joint learning-based method overcomes this drawback and predicts reasonable structure values for the mobile car. It further demonstrates the effectiveness of the joint learning-based model.

\begin{table*}[t!]
	\centering
	\caption{Quantitative results on Cityscapes dataset: Comparison with top performer$^a$}
	\label{tab:Cityscapes}
	\begin{center}
		\resizebox{0.9\linewidth}{!}{%
			\begin{tabular}{|c|c|c|c|c|c|c|c|c|c|}
				\hline
				Method $^b$	&  $\text{W}\times \text{H}$ & \#Params& REL $\downarrow$ & Sq Rel $\downarrow$ &RMSE $\downarrow$ & $\text{RMSE}_{\text{log}} \downarrow$ & $\delta_1 \uparrow$ &  $\delta_2 \uparrow$ & $ \delta_3 \uparrow$ \\ \hline\hline
				Pilzer \textit{et al.} \cite{pilzer2018unsupervised} & $ 512 \times 256 $ & - & 0.240  & 4.264 & 8.049 & 0.334 & 0.710 & 0.871 & 0.937  \\ 
				Struct2Depth 2 \cite{casser2019depth} & $416 \times 128 $ & - & 0.145  &1.737  & 7.280 & 0.205 & 0.813 & 0.942 & 0.976 \\ 
				Monodepth2 \cite{godard2019digging} & $ 416 \times 128 $ & - & 0.129 &1.569 & 6.876 & 0.187& 0.849 & 0.957 & 0.983 \\ 
				ManyDepth \cite{watson2021temporal}  &$416 \times 128$& - & 0.114 &1.193 &6.223& 0.170& 0.875& 0.967& 0.989 \\ 
				\hline
				
				Laina \textit{et al.} \cite{laina2016deeper}  &$416 \times 128$ & - & 0.257 & 4.238 & 7.273 & 0.448 & 0.765 & 0.893 & 0.940 \\ 
				Xu \textit{et al.} \cite{xu2018pad}  &$416 \times 128$ & - & 0.246 & 4.060 & 7.117 & 0.428 & 0.786 & 0.905 & 0.945 \\ 
				Zhang \textit{et al.} \cite{zhang2018joint}  &$416 \times 128$ & 63.6M & 0.234 & 3.776 & 7.104 & 0.416 & 0.776 & 0.903 & 0.949 \\ 
				SDC-Depth \cite{wang2020sdc}  &$416 \times 128$ & 50.4M & 0.227 & 3.800 & 6.917 & 0.414 & 0.801 & 0.913 & 0.950 \\ 
				
				\hline
				
				\textbf{Ours}$^c$ & $ 416 \times 128 $ & 57M & {0.106} & 1.077 & 5.872 & {0.159} & {0.890} & {0.971} &{0.990} \\  
				\textbf{Ours$^c$} & $ 832 \times 416 $ & 57M & \textbf{0.090} & \textbf{0.907} & \textbf{5.518} & \textbf{0.145} & \textbf{0.906} & \textbf{0.980} &\textbf{0.994} \\ 
				\hline
				
		\end{tabular}}
	\end{center}
	\scriptsize
	\begin{tablenotes}
		\item[a]$^a$ The best results of each class are in \textbf{bold}.
		\item[b]$^b$ All methods are classified into self-supervised and supervised learning based.
		\item[c]$^c$ Our model is using ResNeSt-101 encoder and full-shared decoder with $\text{S/D} = 10$.
	\end{tablenotes}
\end{table*}

\subsection{Comparison with state of the art}

\begin{table*}[t!]
	\caption{Quantitative results on class-specific objects: Comparison with top supervised learning methods$^*$.}
	\label{tab:class_specific_state_of_the_art}
	\begin{center}
		\resizebox{0.9\linewidth}{!}{%
			\begin{tabular}{|c|c|c|c|c|c|c|c|c|}
				\hline
				Class	&  Method & REL $\downarrow$ & Sq Rel $\downarrow$ &RMSE $\downarrow$ & $\text{RMSE}_{\text{log}}\downarrow$ & $\delta< 1.25\uparrow$ & $\delta< 1.25^2\uparrow$ & $\delta< 1.25^3\uparrow$ \\ \hline\hline
				\multirow{4}{*}{Street}	
				& DORN & 0.0427 & 0.0630 & 1.2289 & 0.0547 & 0.9935 & 0.9995 & \textbf{0.9999} \\
				& DPT-Hybrid & 0.0435 & 0.0580 & 1.0840 & 0.0521 & 0.9946 & \textbf{0.9997} & \textbf{0.9999} \\
				& Adabins & 0.0400 & 0.0839 & 1.3075 & 0.0527 & 0.9908 & 0.9995 & \textbf{0.9999} \\
				& D+S (Ours) & \textbf{0.0288} & \textbf{0.0443} & \textbf{0.9532} & \textbf{0.0406} & \textbf{0.9947} & \textbf{0.9997} & \textbf{0.9999} \\ 
				\hline
				\multirow{4}{*}{Fence}	
				& DORN & 0.1597 & 1.1038 & 4.5997 & 0.1761 & 0.8077 & 0.9528 & 0.9818 \\
				& DPT-Hybrid & 0.1212 & 0.8112 & 4.0834 & 0.1412 & 0.8579 & \textbf{0.9765} & \textbf{0.9966} \\
				& Adabins & 0.1328 & 1.0730 & 4.6727 & 0.1559 & 0.8336 & 0.9656 & 0.9925 \\
				& D+S (Ours) & \textbf{0.1080} & \textbf{0.7651} & \textbf{3.7522} & \textbf{0.1325} & \textbf{0.8601} & {0.9678} & {0.9958} \\ 
				\hline
				\multirow{4}{*}{Thin Objects} 
				& DORN & 0.2353 & 2.3349 & 6.6373 & 0.2410 & 0.6956 & 0.8983 & 0.9633 \\
				& DPT-Hybrid & 0.1226 & 1.0124 & 5.1132 & 0.1501 & 0.8465 & 0.9674 & 0.9935 \\
				& Adabins & 0.1238 & \textbf{0.9644} & \textbf{4.9617} & 0.1507 & \textbf{0.8510} & \textbf{0.9705} & 0.9940 \\
				& D+S (Ours) & \textbf{0.1190} & {1.0068} & {5.0549} & \textbf{0.1500} & {0.8488} & {0.9657} & \textbf{0.9945} \\ 
				\hline
				\multirow{4}{*}{Person + Rider}	
				& DORN & 0.1794 & 1.5651 & 5.6215 & 0.1926 & 0.7456 & 0.9424 & 0.9846 \\
				& DPT-Hybrid & 0.1226  & \textbf{0.9321}  & \textbf{4.5687} & \textbf{0.1401} & 0.8455 & \textbf{0.9784} & \textbf{0.9984} \\
				& Adabins & 0.1412 & 1.3413 & 5.1942 & 0.1592 & 0.8180 & 0.9631 & 0.9950 \\
				& D+S (Ours) & \textbf{0.1211} & {1.0314} & {4.6541} & {0.1476} & \textbf{0.8462} & {0.9614} & {0.9959} \\ 
				\hline
				\multirow{4}{*}{Car} 
				& DORN & 0.1363  & 0.9280 & 4.2841 & 0.1760 & 0.8767 & 0.9550 & 0.9774 \\
				& DPT-Hybrid & 0.0822 & 0.4605 & 3.4645 & 0.1096 & 0.9378 & \textbf{0.9898} & \textbf{0.9958} \\
				& Adabins & 0.0887  & 0.6665 & 3.9104 & 0.1215 & 0.9283 & 0.9810 & 0.9944 \\
				& D+S (Ours) & \textbf{0.0685} & \textbf{0.4502} & \textbf{3.2816} & \textbf{0.1028} & \textbf{0.9448} & {0.9886}& {0.9949} \\ 
				\hline
				\multirow{4}{*}{Edges of Car}	
				& DORN & 0.1735  & 1.5859 & 5.9826 & 0.2164 & 0.8055 & 0.9320 & 0.9730 \\
				& DPT-Hybrid & 0.0945 & 0.6552 & 4.5450 & 0.1307 & 0.9122 & \textbf{0.9834} & \textbf{0.9941} \\
				& Adabins & 0.1123  & 1.0627 & 5.2884 & 0.1537 & 0.8882 & 0.9685 & 0.9902 \\
				& D+S (Ours) & \textbf{0.0825} & \textbf{0.6536} & \textbf{4.3600} & \textbf{0.1262} & \textbf{0.9197} & {0.9823}& {0.9929} \\
				\hline
				
		\end{tabular}}
	\end{center}
	\scriptsize
	\begin{tablenotes}
        \item[a]$^*$ The evaluation metrics are based on the \textbf{KITTI Eigen Zhou test} with improved ground-truth. The best results of each class are in \textbf{bold}. The training image resolution for all models is $1024 \times 320$. All compared models are pretrained and fine-tuned on KITTI dataset. Ours D+S model is the using ResNeSt 101 encoder, full-shared decoder, S/D = 10 and pretrained on Cityscape dataset. 
        
    \end{tablenotes}
\end{table*}

To demonstrate the efficiency and robustness, we evaluate the proposed model on two challenging datasets in the field of depth estimation,  i.e., KITTI \cite{geiger2012we} and Cityscapes \cite{Cordts2016Cityscapes}. First, several results of quantitative comparisons with learning-based state-of-the-art methods on KITTI \cite{geiger2012we} and Cityscapes \cite{Cordts2016Cityscapes} are shown in Table \ref{tab:state_of_the_art} and \ref{tab:Cityscapes}, respectively. On the KITTI dataset, our pretrained model that is pretrained on Cityscapes dataset, has the best performance in terms of relative error and similar performance of threshold ($\sigma$) compared to state-of-the-art depth supervised learning methods and self-supervised learning methods (See Table~\ref{tab:state_of_the_art}).
In order to evaluate the influence of proposed components on final prediction, we also conduct the experiments with different encoders and structure ground-truth generated by different P+P pipelines. In the experiments, three combinations are compared, namely \textit{VCN} P+P with \textit{ResNeSt101}, \textit{Rigidmask} P+P with \textit{ResNet101}, and \textit{Rigidmask} P+P with \textit{ResNeSt101}. 
From Table \ref{tab:state_of_the_art}, it can be seen that P+P geometry has the greatest impact on the final performance, e.g., with REL improving by 0.018 via \textit{Rigidmask} P+P pipeline and only 0.005 via \textit{ResNeSt101}.
From Table~\ref{tab:Cityscapes}, it is easy to see that our model achieves the best performance for all of the metrics on the Cityscapes dataset. Note that, it is unfair to compare a supervised learning approach with a self-supervised learning method, however, considering recent explosive development in the field of self-supervision, we think it is necessary to make a comparison of accuracy without training time here. Moreover, most of the depth supervised learning cannot predict the area, where the ground-truth is empty and the self-supervised learning method can estimate it well. Our model combines both the precision of the supervised method and the inferring ability of the self-supervised learning approach, which can be further evident in qualitative results.

\begin{figure*}[t!]
        \begin{center}
        \resizebox{0.9\linewidth}{!}{%
		\addtolength{\tabcolsep}{-5pt}    
		\renewcommand{\arraystretch}{0.1}
		
		\begin{tabular}{lcccc}
			Input&
			\includegraphics[width=0.2\textwidth]{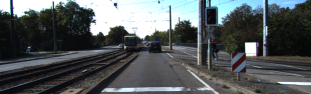}&
			\includegraphics[width=0.2\textwidth]{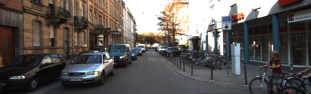}&
			\includegraphics[width=0.2\textwidth]{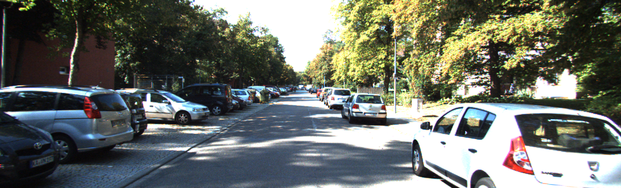}&
			\includegraphics[width=0.2\textwidth]{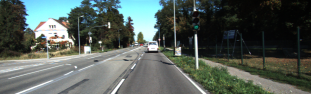}
			\\
			Ground-Truth &
			\includegraphics[width=0.2\textwidth]{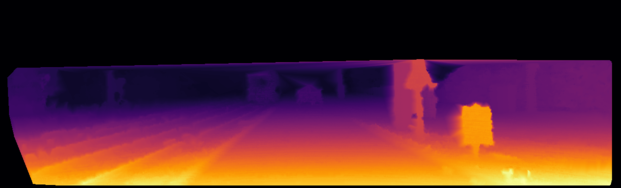}&
			\includegraphics[width=0.2\textwidth]{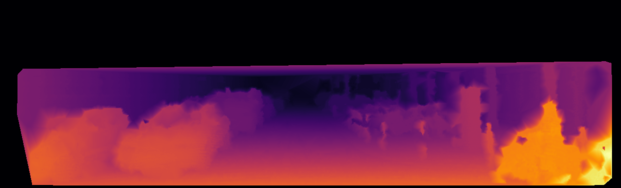}&
			\includegraphics[width=0.2\textwidth]{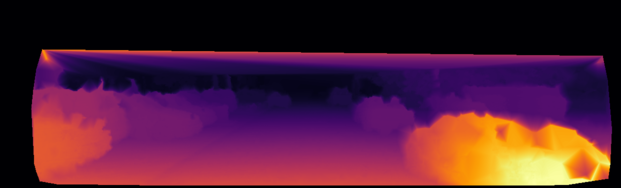}&
			\includegraphics[width=0.2\textwidth]{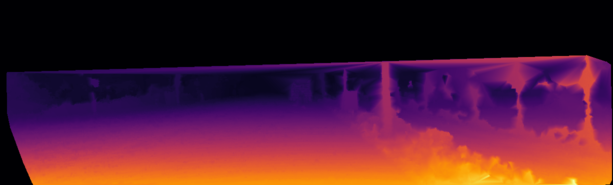}\vspace{0.07cm}
            \\ \hline 
			SfMLearner \cite{zhou2017unsupervised}&
			\includegraphics[width=0.2\textwidth]{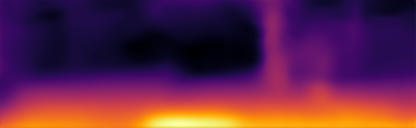}&
			\includegraphics[width=0.2\textwidth]{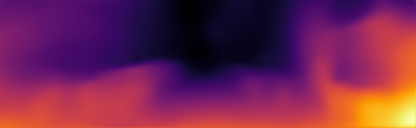}&
			\includegraphics[width=0.2\textwidth]{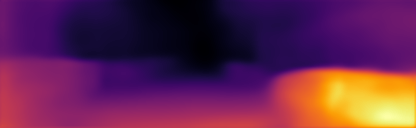}&
			\includegraphics[width=0.2\textwidth]{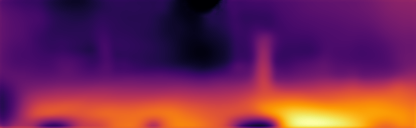}
			\\
			GeoNet \cite{yin2018geonet}&
			\includegraphics[width=0.2\textwidth]{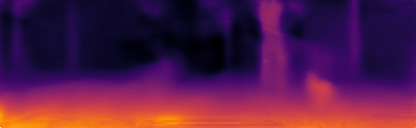}&
			\includegraphics[width=0.2\textwidth]{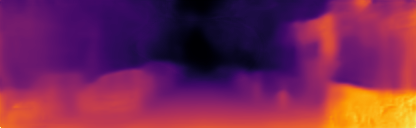}&
			\includegraphics[width=0.2\textwidth]{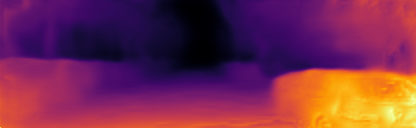}&
			\includegraphics[width=0.2\textwidth]{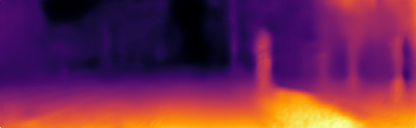}
			\\
			DDVO \cite{wang2018learning}&
			\includegraphics[width=0.2\textwidth]{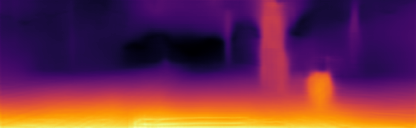}&
			\includegraphics[width=0.2\textwidth]{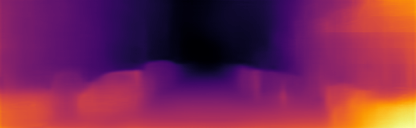}&
			\includegraphics[width=0.2\textwidth]{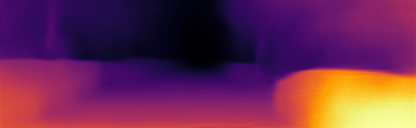}&
			\includegraphics[width=0.2\textwidth]{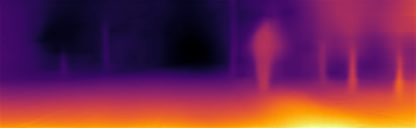}
			\\
			Monodepth2 \cite{godard2019digging}&
			\includegraphics[width=0.2\textwidth]{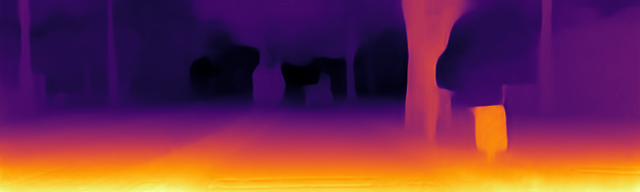}&
			\includegraphics[width=0.2\textwidth]{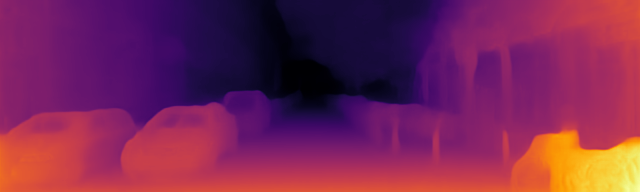}&
			\includegraphics[width=0.2\textwidth]{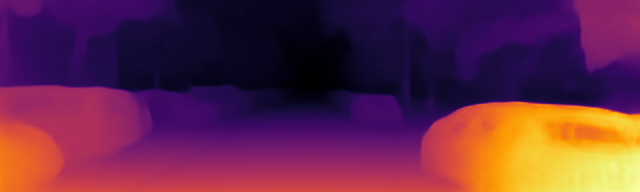}&
			\includegraphics[width=0.2\textwidth]{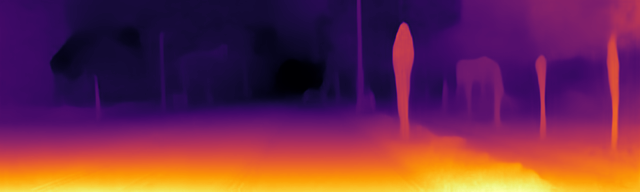}
			\\ \hline 

			\multirow{2}{*}{ResNet101} &
			\includegraphics[width=0.2\textwidth]{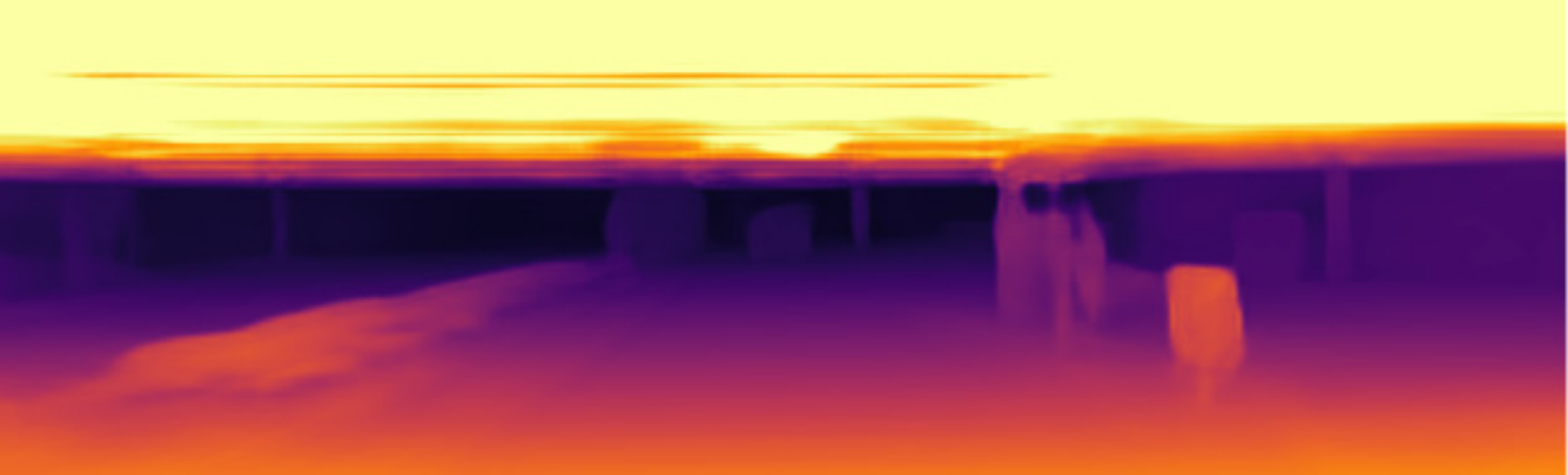}&
			\includegraphics[width=0.2\textwidth]{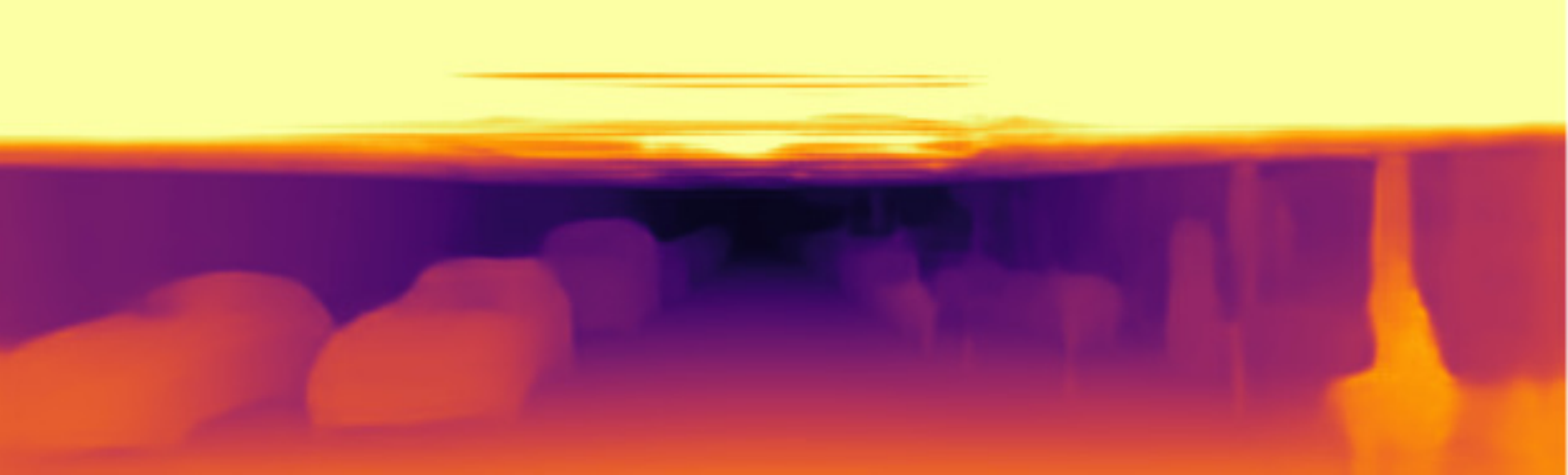}&
			\includegraphics[width=0.2\textwidth]{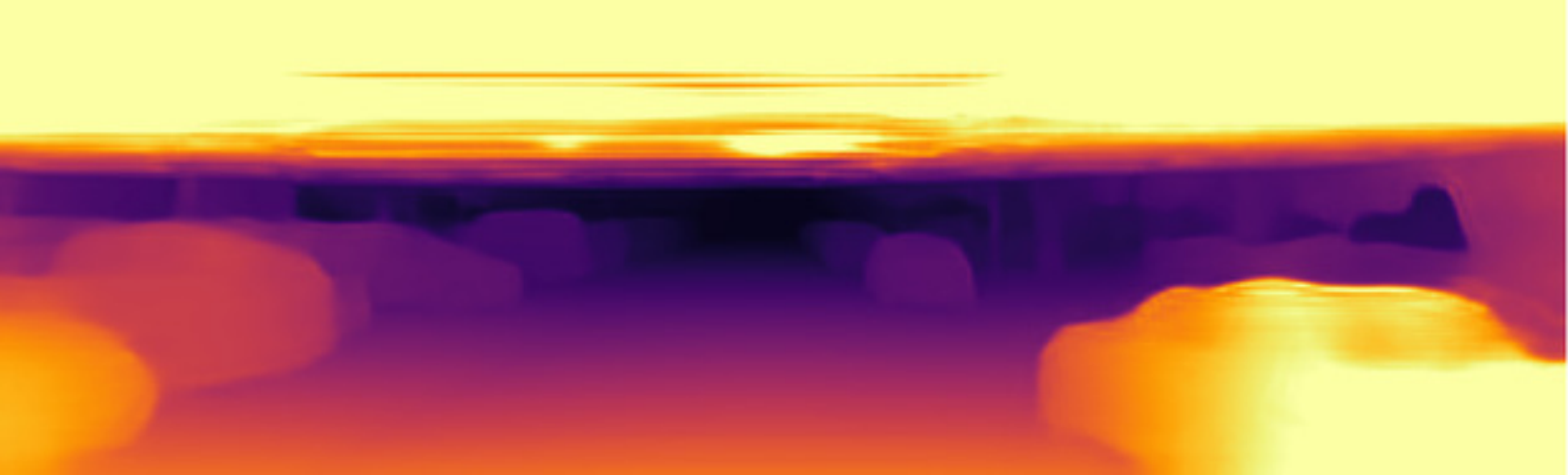}&
			\includegraphics[width=0.2\textwidth]{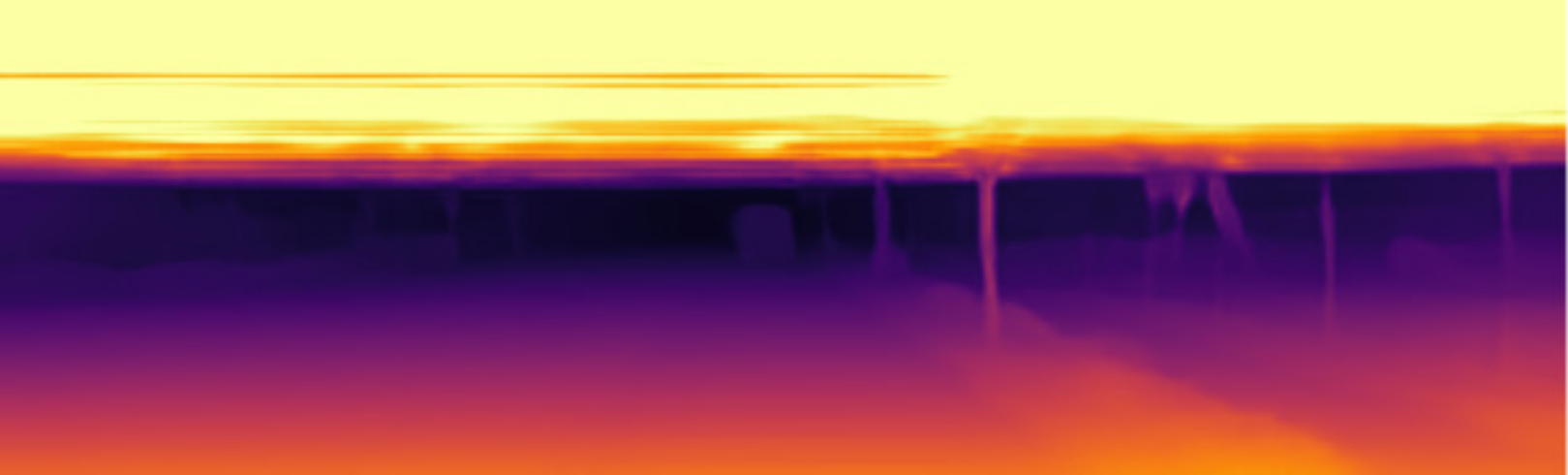}
            \\
            &
			\includegraphics[width=0.2\textwidth]{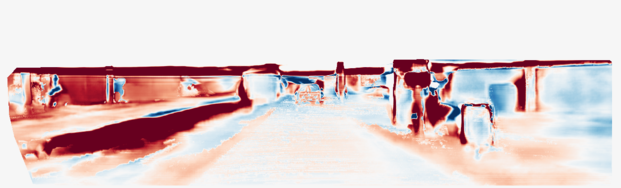}&
			\includegraphics[width=0.2\textwidth]{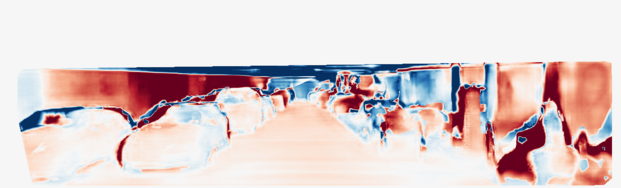}&
			\includegraphics[width=0.2\textwidth]{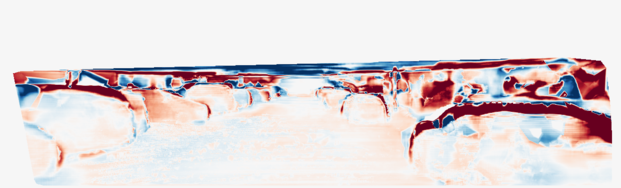}&
			\includegraphics[width=0.2\textwidth]{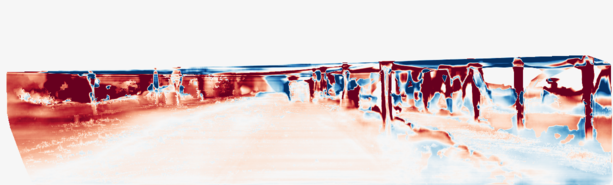}\vspace{0.07cm}
            \\
            \multirow{2}{*}{DORN \cite{fu2018deep}} &
			\includegraphics[width=0.2\textwidth]{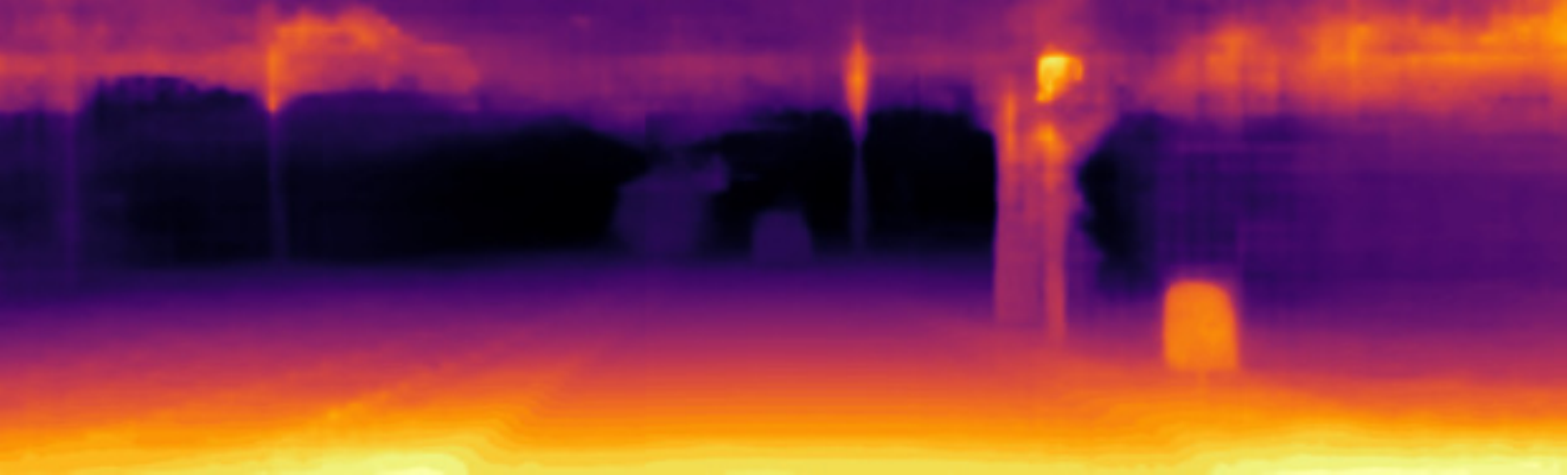}&
			\includegraphics[width=0.2\textwidth]{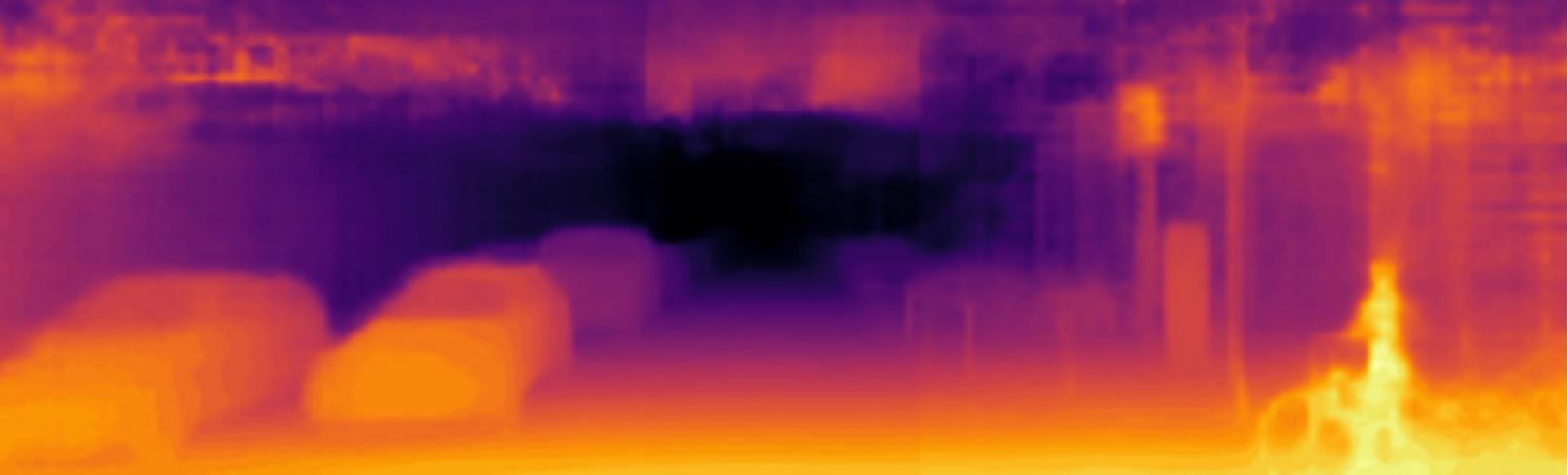}&
			\includegraphics[width=0.2\textwidth]{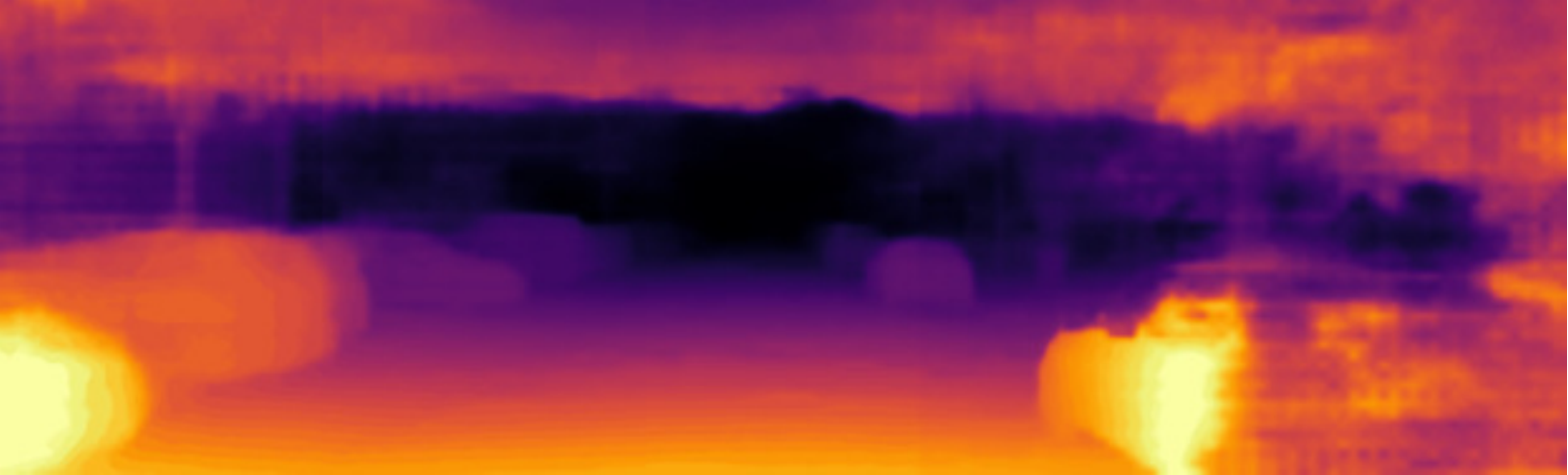}&
			\includegraphics[width=0.2\textwidth]{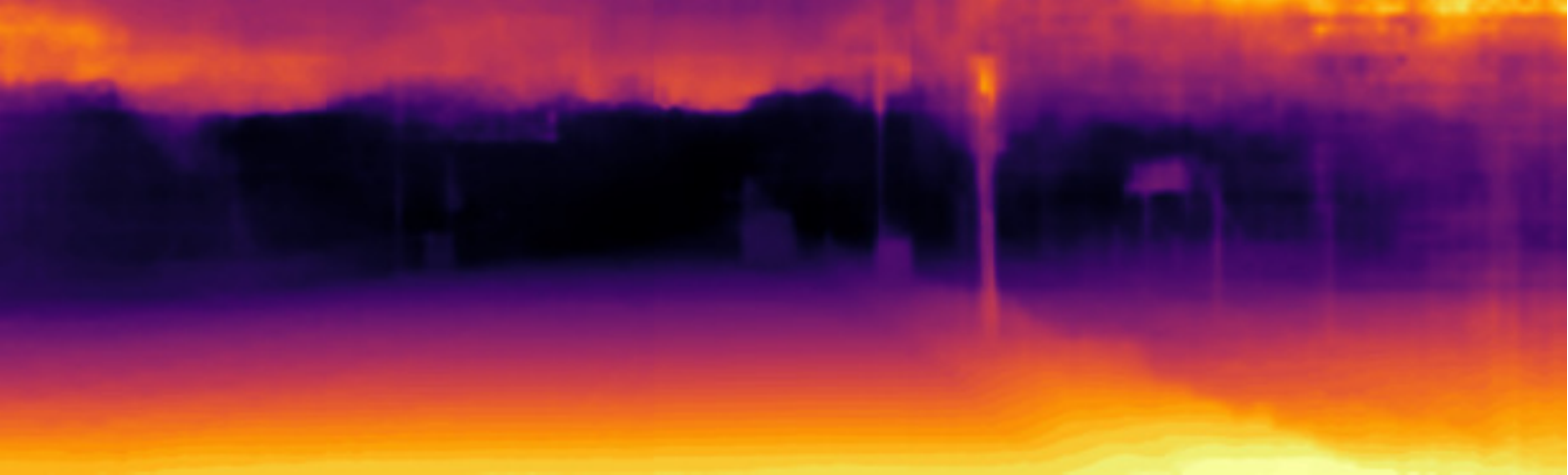}\vspace{0.07cm}
            \\
            &
			\includegraphics[width=0.2\textwidth]{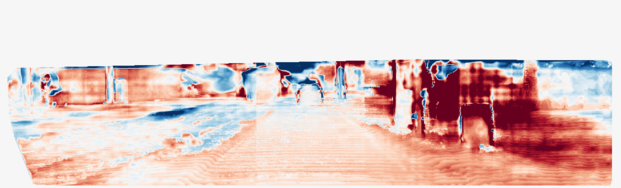}&
			\includegraphics[width=0.2\textwidth]{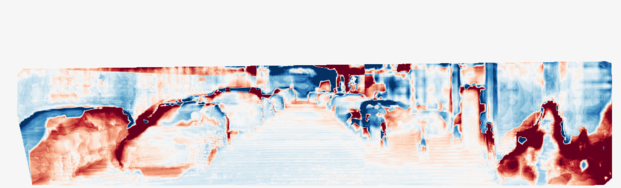}&
			\includegraphics[width=0.2\textwidth]{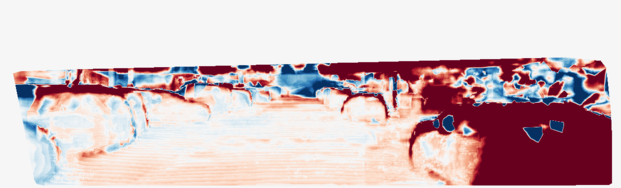}&
			\includegraphics[width=0.2\textwidth]{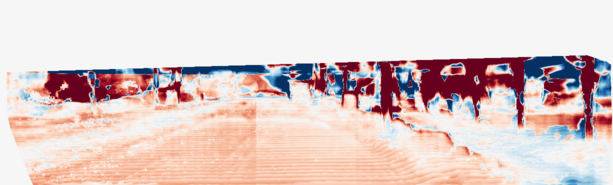}\vspace{0.07cm}
            \\
            \multirow{2}{*}{DPT-Hybrid \cite{ranftl2021vision}} &
			\includegraphics[width=0.2\textwidth]{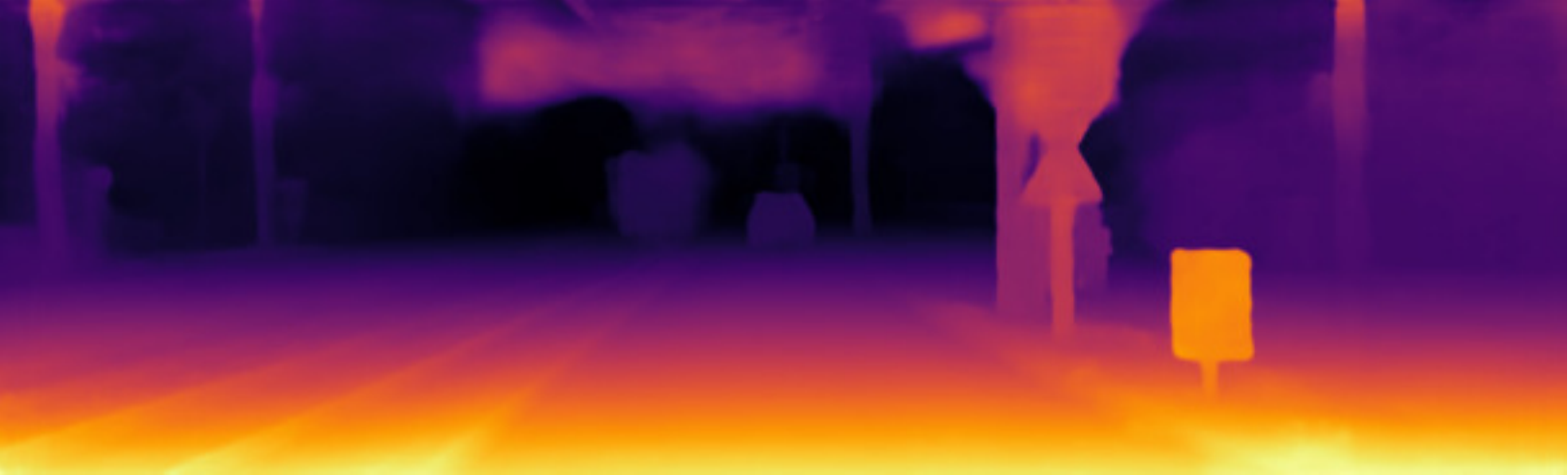}&
			\includegraphics[width=0.2\textwidth]{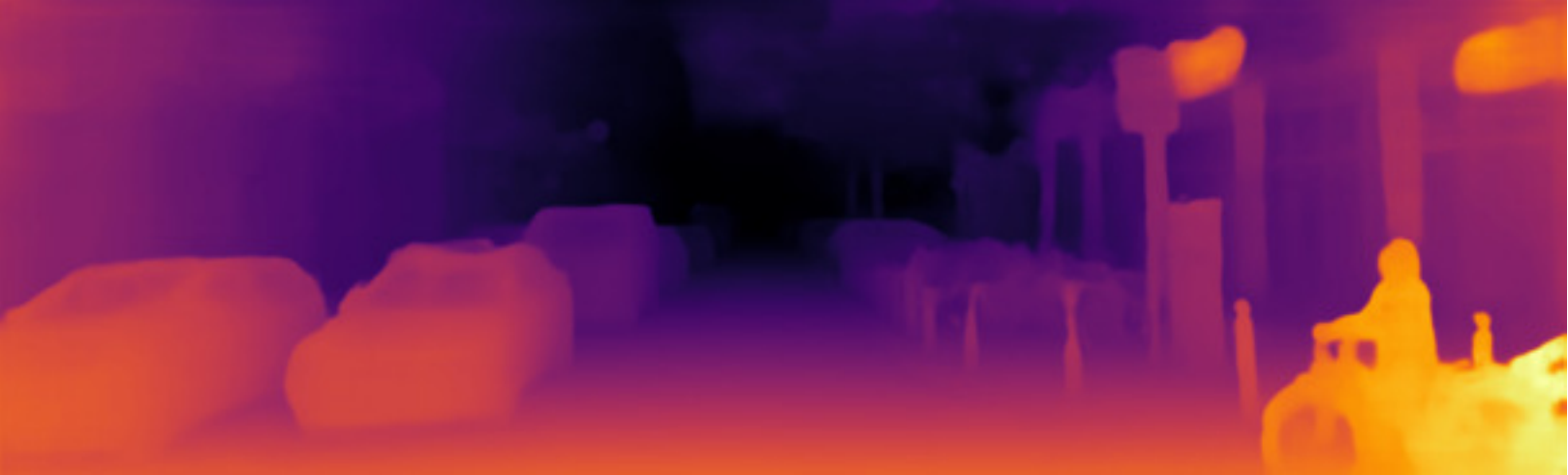}&
			\includegraphics[width=0.2\textwidth]{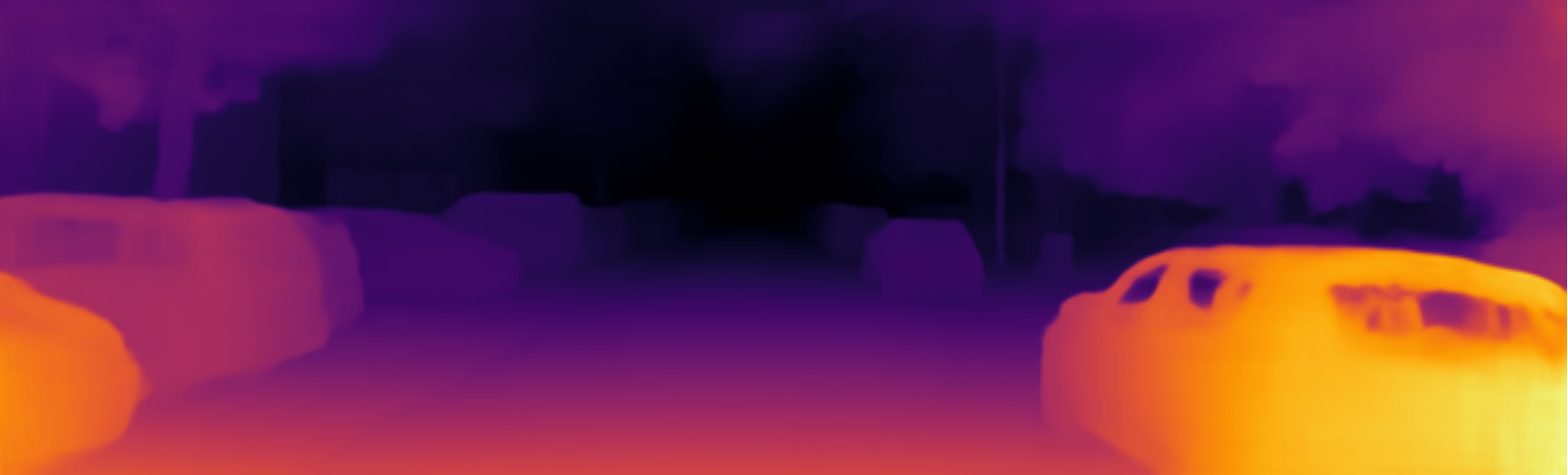}&
			\includegraphics[width=0.2\textwidth]{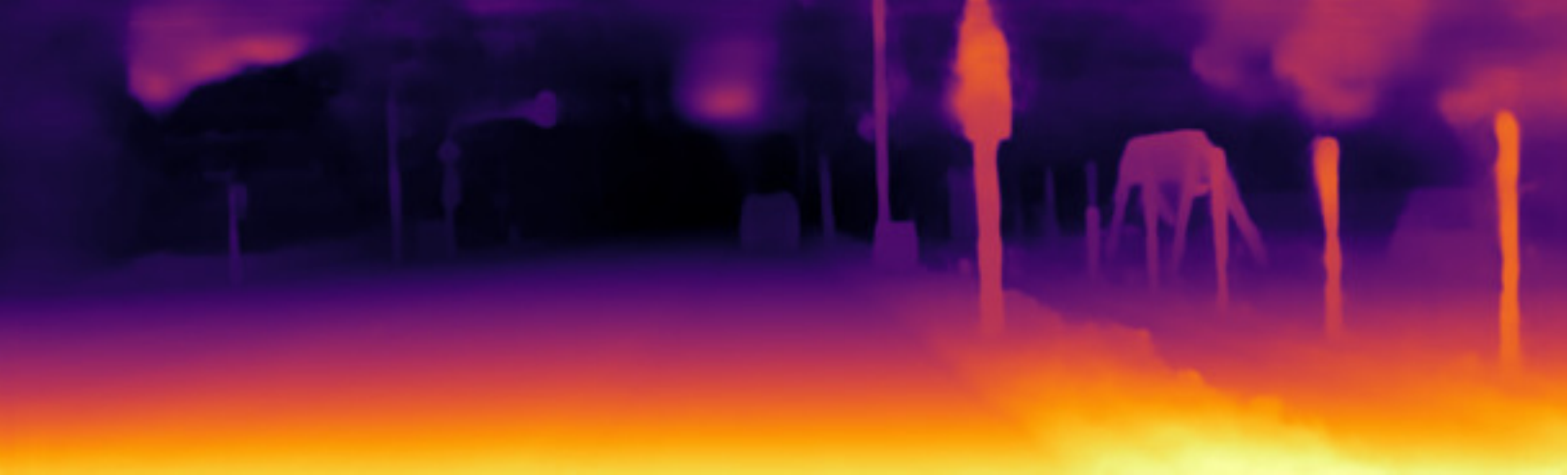}\vspace{0.07cm}
            \\
            &
			\includegraphics[width=0.2\textwidth]{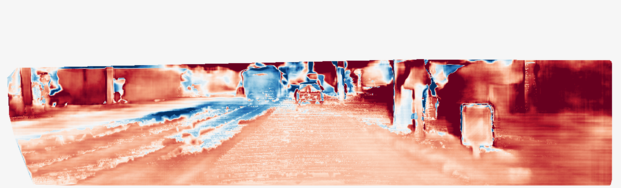}&
			\includegraphics[width=0.2\textwidth]{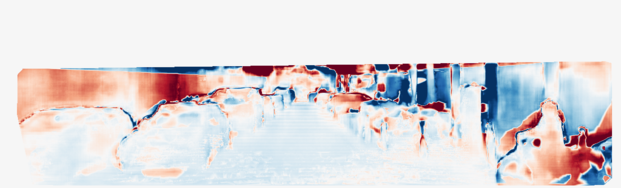}&
			\includegraphics[width=0.2\textwidth]{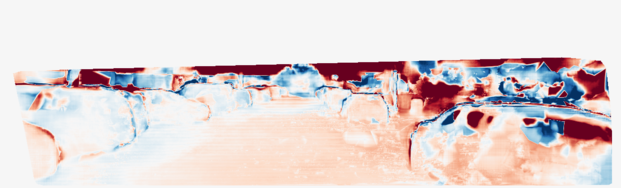}&
			\includegraphics[width=0.2\textwidth]{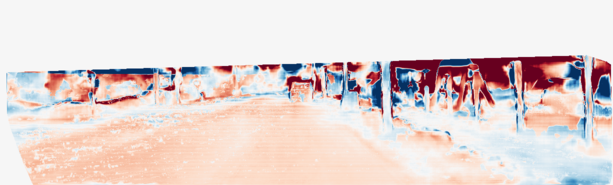}\vspace{0.07cm}
            \\
            \multirow{2}{*}{Adabins \cite{bhat2021adabins}}  &
			\includegraphics[width=0.2\textwidth]{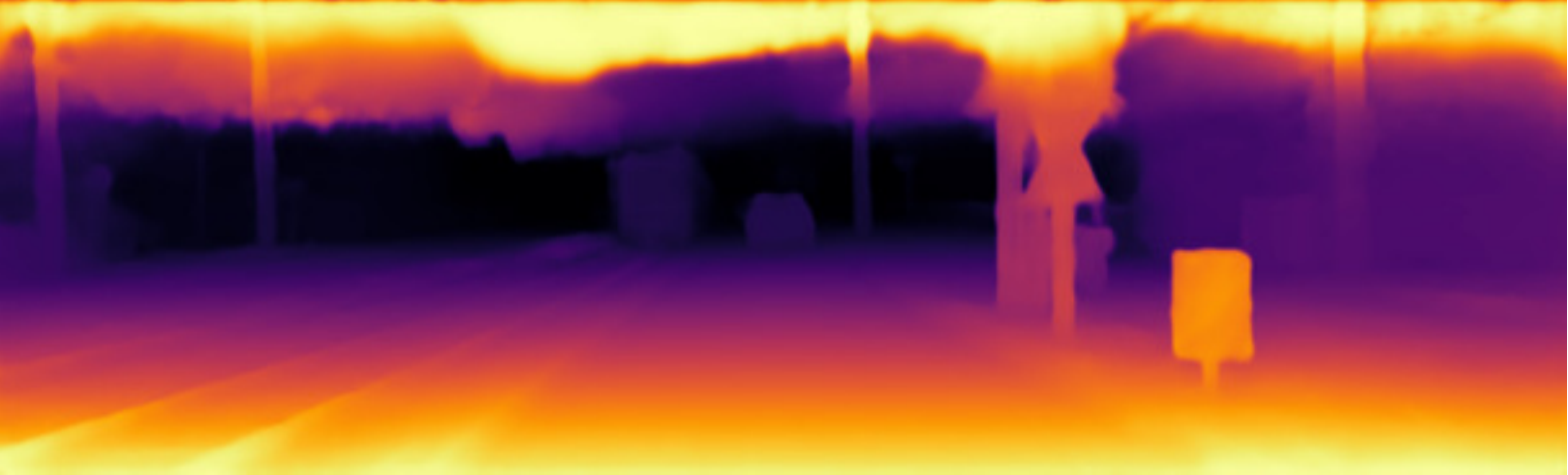}&
			\includegraphics[width=0.2\textwidth]{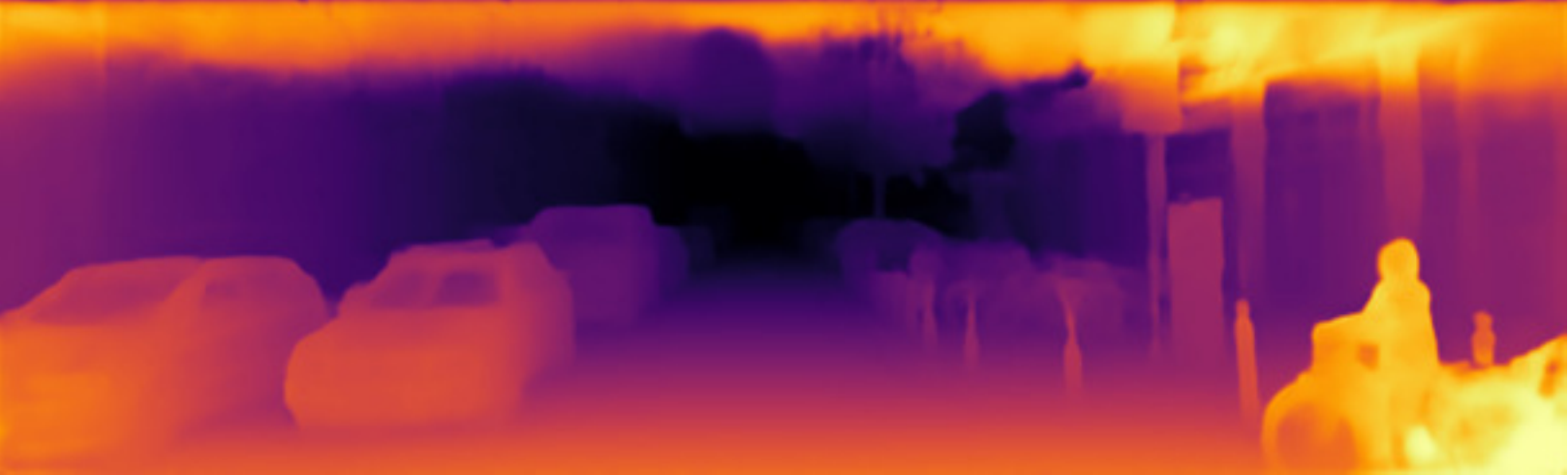}&
			\includegraphics[width=0.2\textwidth]{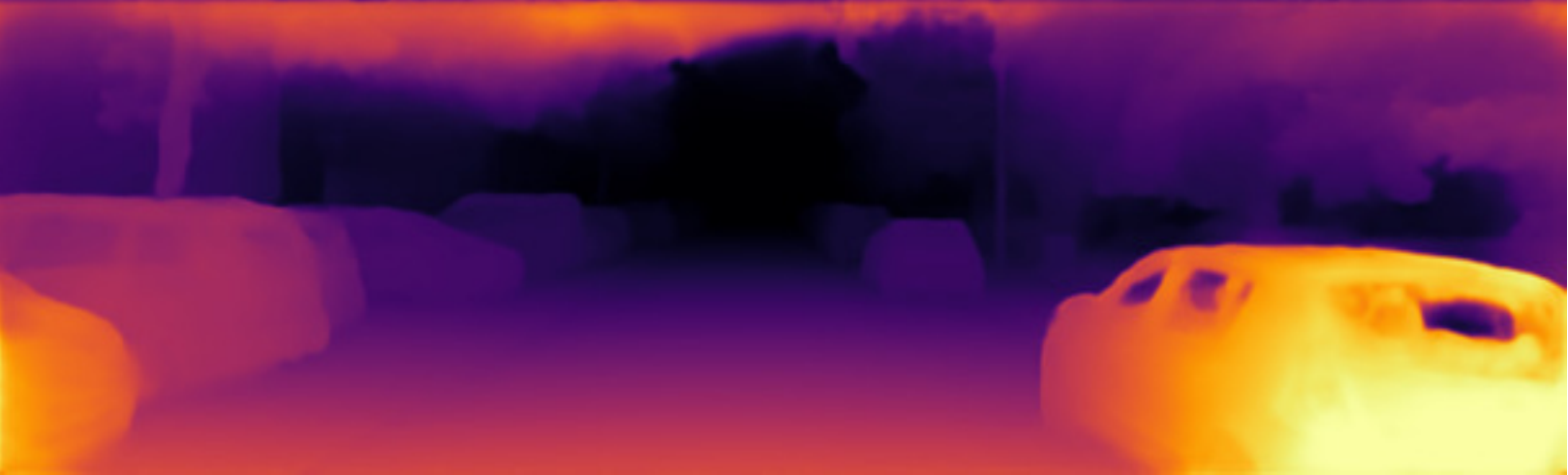}&
			\includegraphics[width=0.2\textwidth]{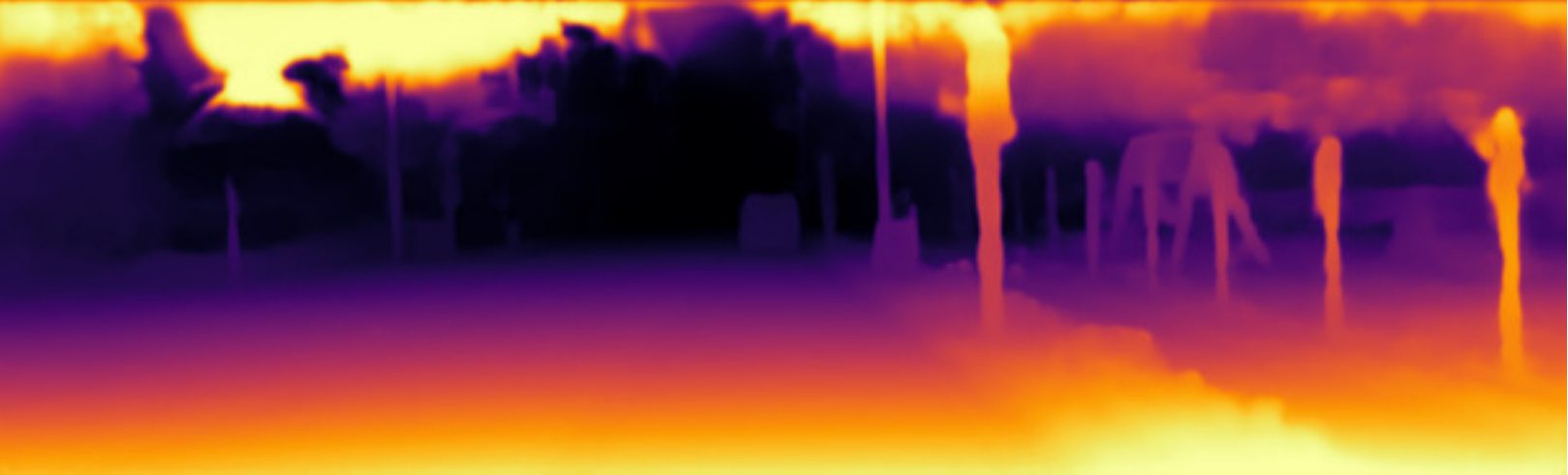}
            \\
            &
			\includegraphics[width=0.2\textwidth]{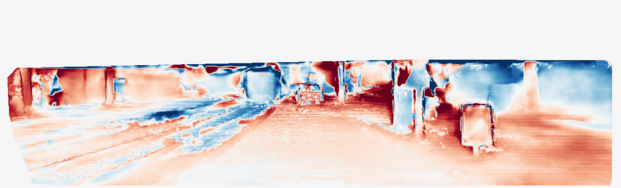}&
			\includegraphics[width=0.2\textwidth]{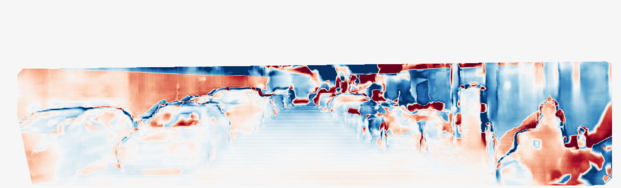}&
			\includegraphics[width=0.2\textwidth]{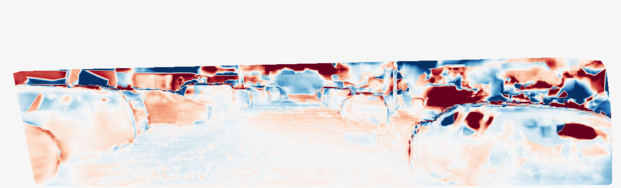}&
			\includegraphics[width=0.2\textwidth]{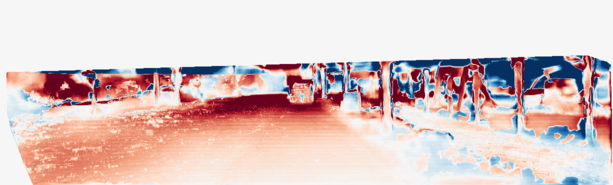}\vspace{0.07cm}
            \\
			\textbf{D+S (ours)}&
			\includegraphics[width=0.2\textwidth]{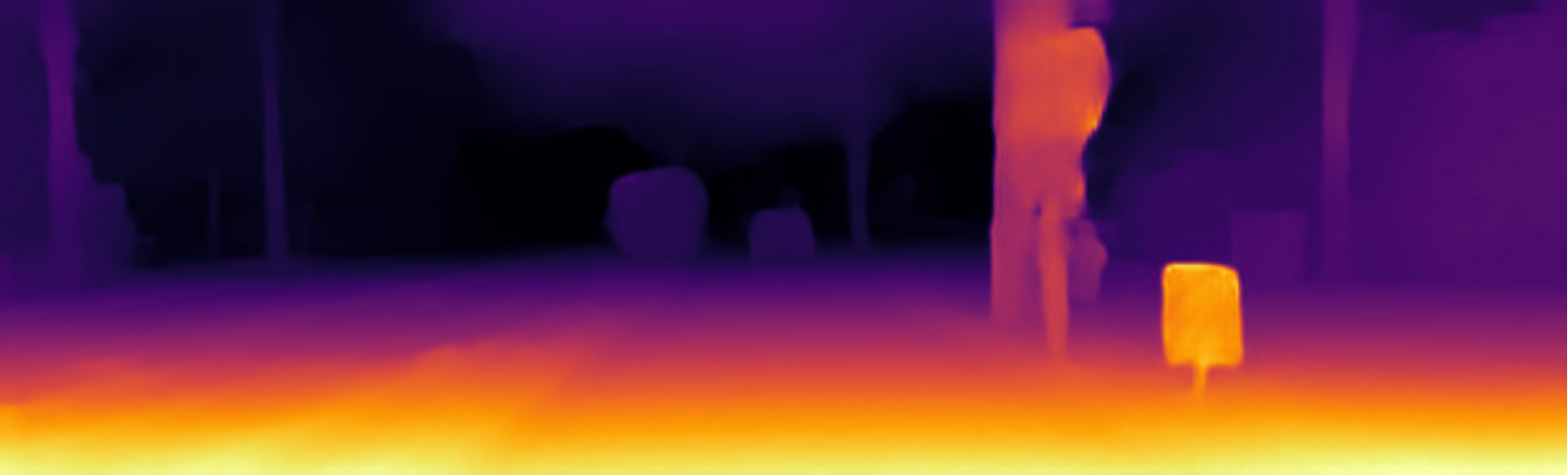}&
			\includegraphics[width=0.2\textwidth]{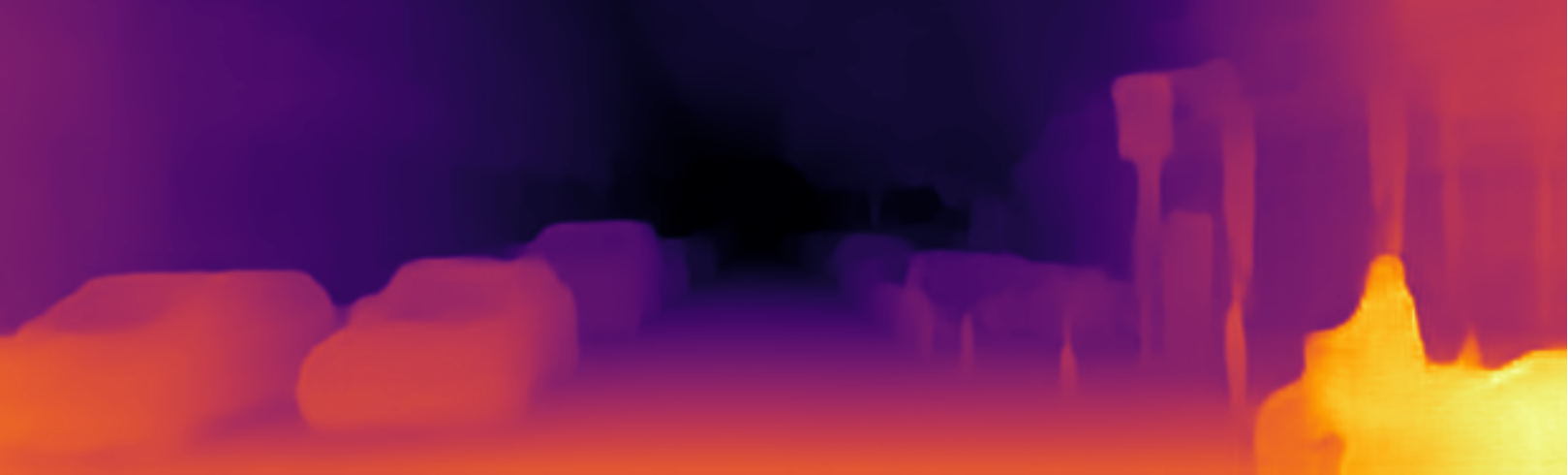}&
			\includegraphics[width=0.2\textwidth]{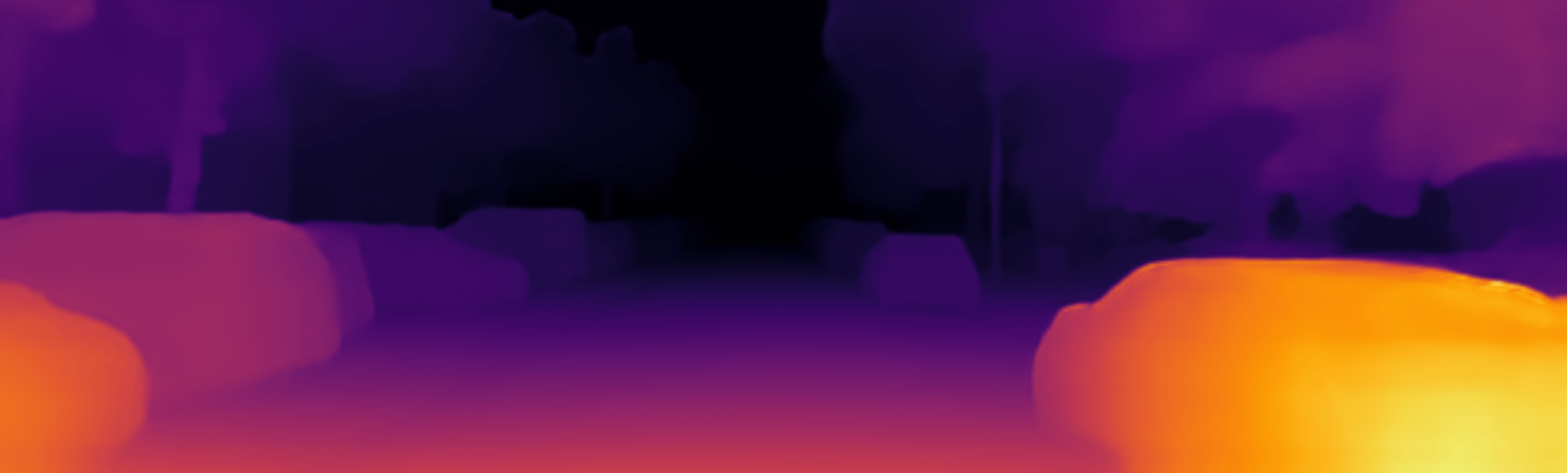}&
			\includegraphics[width=0.2\textwidth]{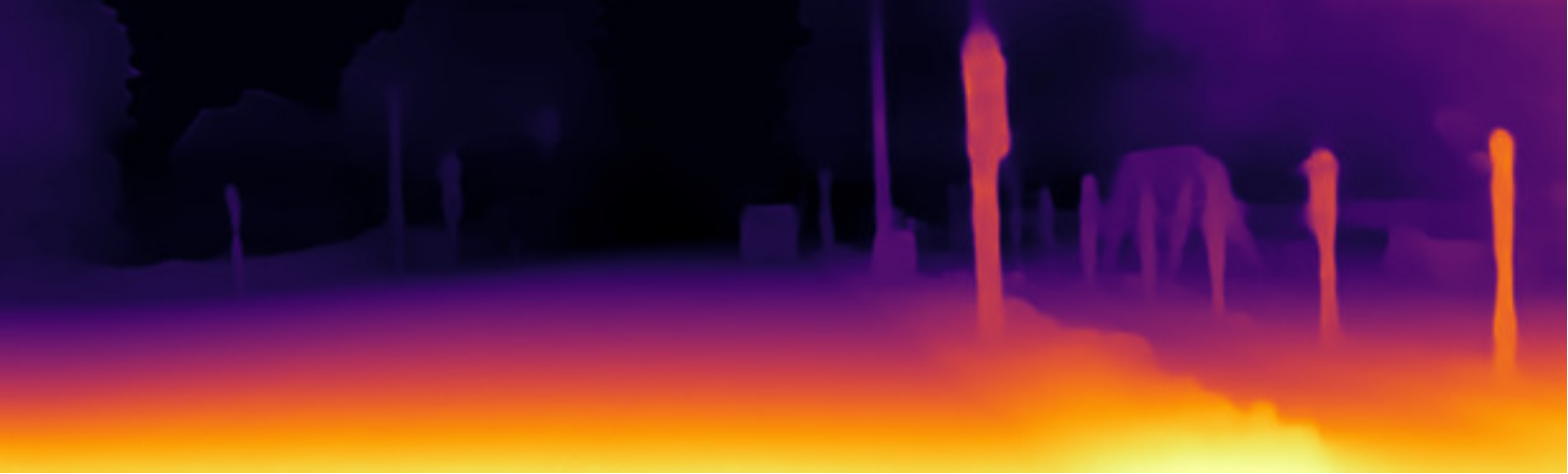}

		\end{tabular}}
		\caption{\textbf{Qualitative comparison with popular exiting methods:} Row 1-2: Original RGB images from test set and improved \emph{ground-truth} depth (disparity) images; Each subsequent row: depth map generated by a mono-based method. Ours method is using \textit{ResNeSt 101} as encoder and has one shared decoder with S/D $=10$. In comparisons with other supervised learning methods, we list the \emph{REL} differences between our \emph{D+S} method and the compared approach below each depth prediction, where red region indicates our model is better and blue area is opposite.}
		\label{fig:state_of_the_art} %
		\end{center}
\end{figure*}

In addition to the comparison of depth prediction over global image region, we also compare predictions for specific regions where thin objects and edges are located with other top supervised methods. As demonstrated in Table~\ref{tab:class_specific_state_of_the_art}, our Depth+Structure model performs better in terms of \textit{REL}. Additionally, the results of the class ``Fence'', ``Thin Objects'', and ``Edges of Car'' further demonstrate the effectiveness of the proposed model, i.e., combining the structure information has better representation on thin objects and edges. More evidences can be found on qualitative results.



The qualitative results comparing other good practices are shown in Fig. \ref{fig:state_of_the_art}. 
The impressive achievement of our model is the inferring of empty areas in the ground-truth and the accurate prediction of thin objects. As aforementioned, our model overcomes the drawback of the supervised learning method and predicts the depth of the top region well even in the absence of ground-truth. Compared to both supervised and self-supervised learning methods, our model has higher accuracy on the street, edges and thin objects, such as riders, traffic lamps and cars. Another interesting observation is that all top supervised learning methods ignore the glass, such as car side window glass and windshield (see 3. column of Fig. \ref{fig:state_of_the_art}), and the value at the glass region is the depth of the object after transmission. Since glass is invisible to LiDAR sensors, the depth ground-truth collected by LiDAR lacks information of glass, which causes the supervised learning methods to learn wrong predictions around these areas. From the qualitative results, it is clear that our \emph{D+S} model successfully predicts the depth value of glasses even with lack of ground-truth. This observation further demonstrate the effectiveness of joint learning using the additional structure information.

\section{Conclusions}
{
    
    In this paper, we introduce a complete framework for joint estimation of monocular depth and structure. The framework consists of a joint supervision model, which predicts dense depth and structure map from single RGB image. To this end, we develop a plane and parallax geometry pipeline that offers dense structure ground-truth information for each frame. We coupled the depth regression with a structure supervised learning branch to enable precise depth prediction around thin objects and edges, and inferring depth value in unlabeled areas. Furthermore, we have demonstrated that our joint model cannot only enhance the monocular depth estimation, but also the structure prediction. Using our model, the dense structure information from extreme views of camera under no ego-motion condition are successfully predicted, including the area around epipole. Experiments with a wide variety of outdoor image content at different cities show that our model achieves realistic joint prediction of dense depth and structure from single view.
    
    The study shows that our model can be of high relevance for various use-cases that need scene reconstruction, e.g., it can be used in the field of autonomous driving which requires a more robust and safer representation of a 3D scene. Furthermore, with monocular depth estimation, the cost will be greatly reduced. The predicted structure is another representation of the scene and has been proven to be used to improve the accuracy of monocular depth prediction. However, the structure information is highly relying on the defined reference plane. For an outdoor environment, the reference plane can be selected as the street surface, while in an indoor environment, the ground plane is usually occluded by other objects. Therefore, we plan to break through the limitation and try to implement it on an indoor mobile robotic system.
}

\section*{ACKNOWLEDGMENT}
This work is supported by the funding of the Lighthouse Initiative Geriatronics by StMWi Bayern (Project X, grant no. 5140951) and LongLeif GaPa GmbH (Project Y, grant no. 5140953).

\section{Appendix}
\label{sec:appendix}


	
	\begin{figure}[h]
        \begin{centering}
        \begin{tabular}{c}
            \includegraphics[width=0.3\textwidth]{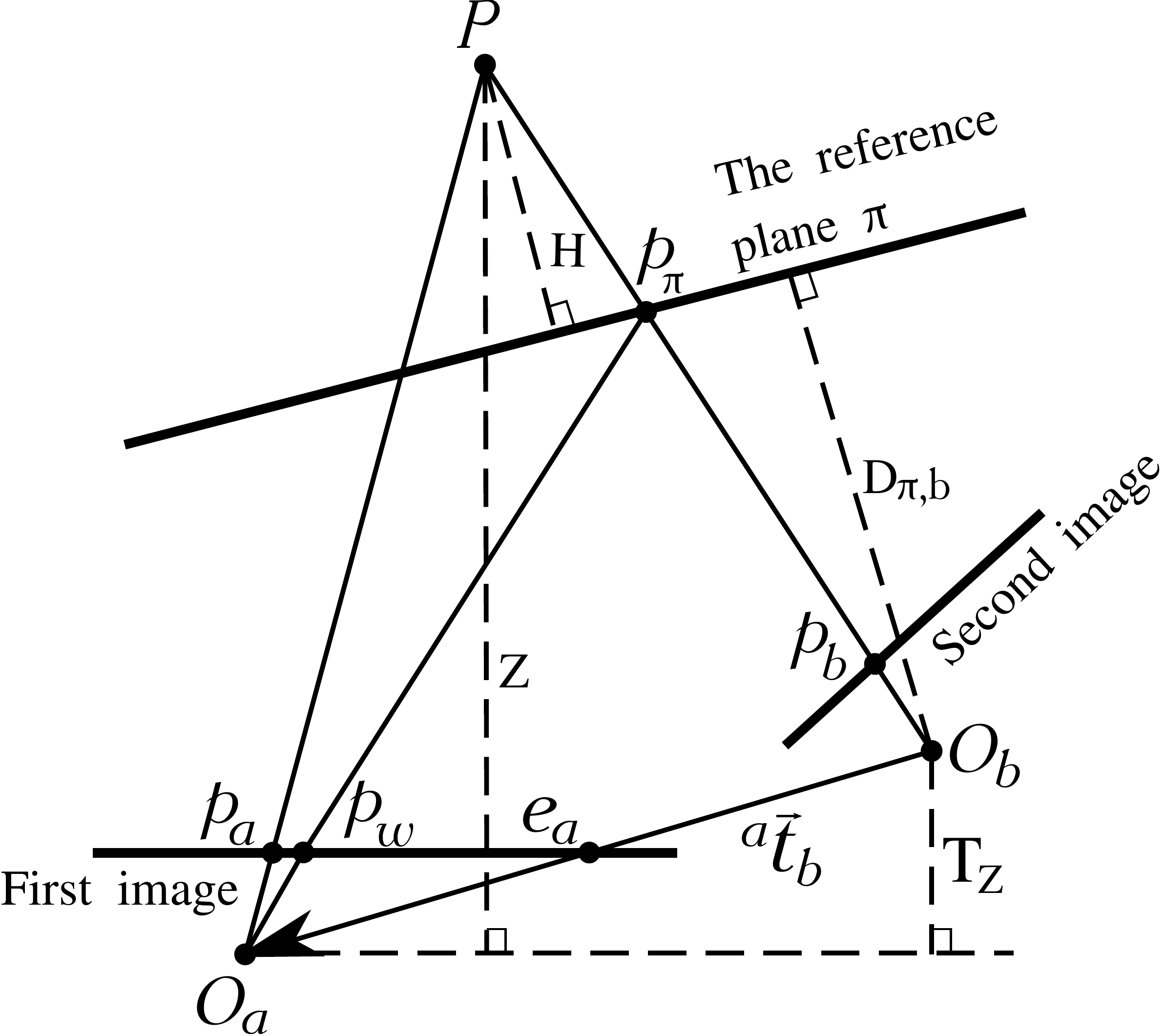}
        \end{tabular}
        \caption{The geometric interpretation of the Plane+Parallax decomposition.
        }
        \label{fig:geometry}
        \end{centering}
    \end{figure}
    

    
As shown in the Fig~\ref{fig:geometry}, the 3D point $P$ in two camera coordinates can be expressed as:
\begin{subequations}
\begin{align}
    \label{eq:7a}
    \vec{P}_b &= {^b\mathbf{R}_a}\vec{P}_a + {^b\vec{t}_a} \\
    \label{eq:7b}
    \vec{P}_a &= {^a\mathbf{R}_b}\vec{P}_b + {^a\vec{t}_b} 
\end{align}
\end{subequations}
where $\mathbf{R}$ and $\vec{t}$ denote the rotation matrix and translation vector between two camera coordinates. 

For the reference plane $\pi$ in the Fig~\ref{fig:geometry}, it can be a normal vector $\vec{n}_i$ and its orthogonal distance $D_{\pi, i}$ to the coordinate $i$ origin, where the coordinate frame is one of the cameras ($i\in\{a,b\}$). Hence, the relation between any 3D point $P$ in the coordinate and the reference surface can be represented by:
\begin{equation}
    \label{eq:8}
    \vec{n}^T_i\vec{P}_i = D_{\pi, i}+H
\end{equation}
where $H$ is the distance from the 3D point $P$ to the reference plane $\pi$. And the Eq~\ref{eq:8} can be reformed as $1 = (\vec{n}^T_i\vec{P}_i-H)/{D_{\pi, i}}$, which is substituted into Eq~\ref{eq:7b} (or Eq~\ref{eq:7a}) as following:
\begin{equation}
\begin{aligned}
    \label{eq:9}
    \vec{P}_a &= {^a\mathbf{R}_b}\vec{P}_b + {^a\vec{t}_b} \frac{\vec{n}^T_b\vec{P}_b-H}{D_{\pi, b}} \\
    &= \left({^a\mathbf{R}_b}+\frac{{^a\vec{t}_b} \vec{n}^T_b}{D_{\pi, b}} \right)\vec{P}_b - \frac{H}{D_{\pi, b}}{^a\vec{t}_b}
\end{aligned}
\end{equation}

Assume that the camera $a$ and $b$ are the same camera at different places and the camera intrinsic parameter matrix $\mathbf{K}$ are not changing during movement. Then the 3D point $\vec{P}_i$ can be expressed by the projection pixel as $\vec{P}_i =\mathbf{K}^{-1}{Z_i}p_i$, where $Z_i$ is the depth value of the 3D point $\vec{P}$ at the camera frame $i$. When substituting the equation into the Eq~\ref{eq:9}, it results in:

\begin{subequations}
\begin{align}
    \label{eq:10a}
    \frac{Z_a}{Z_b}\vec{p}_a &= \mathbf{K}\left( {^a\mathbf{R}_b} +\frac{{^a\vec{t}_b} \vec{n}^T_b}{D_{\pi, b}}\right)\mathbf{K}^{-1}\vec{p}_b - \frac{H}{D_{\pi, b}Z_b}\mathbf{K}{^a\vec{t}_b} \\
    \label{eq:10b}
    \vec{p}_a&\simeq{^a\mathbf{A}_b}\vec{p}_b -  \frac{H}{D_{\pi, b}Z_b}{\vec{T}}
\end{align}
\end{subequations}

where ${Z_a}/{Z_b}$ is an arbitrary scale, $\simeq$ denotes equality to up an arbitrary scale, $^a\mathbf{A}_b = \mathbf{K}\left( {^a\mathbf{R}_b} +({{^a\vec{t}_b} \vec{n}^T_b})/{D_{\pi, b}}\right)\mathbf{K}^{-1}$ is the homography matrix between two camera frames due to the reference plane $\pi$, and $\vec{T} = \mathbf{K}{^a\vec{t}_b}$. Hence, the projection process can be represented by scaling both sides by their third component of homography matrix as follows:
\begin{equation}
\begin{aligned}
\label{eq:11}
        \vec{p}_a &= \frac{{^a\mathbf{A}_b}\vec{p}_b -  \frac{H}{D_{\pi,b}Z_b}\vec{T}} {{\vec{a}^T_{3}}\vec{p}_b -  \frac{H}{D_{\pi, b}Z_b}{T_{_Z}}} \\
        &= \frac{{^a\mathbf{A}_b}\vec{p}_b} {{\vec{a}^T_{3}}\vec{p}_b} - \frac{{^a\mathbf{A}_b}\vec{p}_b}{{\vec{a}^T_{3}}\vec{p}_b} + \frac{{^a\mathbf{A}_b}\vec{p}_b -  \frac{H}{D_{\pi, b}Z_b}\vec{T}}{{\vec{a}^T_{3}}\vec{p}_b -  \frac{H}{D_{\pi, b}Z_b}{T_{_Z}}} \\
        & = \frac{{^a\mathbf{A}_b}\vec{p}_b} {{\vec{a}^T_{3}}\vec{p}_b} + \frac{{^a\mathbf{A}_b}\vec{p}_b} {{\vec{a}^T_{3}}\vec{p}_b}\frac{\frac{H T_{_Z}}{D_{\pi, b}Z_b}}{{\vec{a}^T_{3}}\vec{p}_b - \frac{H T_{_Z}}{D_{\pi, b}Z_b}} \\
        &\ \ \ - \frac{\frac{H}{D_{\pi, b}Z_b}}{{\vec{a}^T_{3}}\vec{p}_b - \frac{H T_{_Z}}{D_{\pi, b}Z_b}}\vec{T}
\end{aligned}
\end{equation}

where ${\vec{a}^T_{3}}$ is the third row vector of ${^a\mathbf{A}_b}$ and $T_{_Z}$ is the third element in $\vec{T}$. Note that the third component of $\vec{p}_a$ is $1$. Considering the arbitrary scale in the Eq~\ref{eq:10a}, it can be obtained as:
\begin{equation}
\label{eq:12}
    \frac{Z_a}{Z_b} = {\vec{a}^T_{3}}\vec{p}_b - \frac{H T_{_Z}}{D_{\pi, b}Z_b}
\end{equation}
Substituting the Eq~\ref{eq:12} into the Eq~\ref{eq:11}, the pixel position can be simplified as:
\begin{equation}
\label{eq:13}
    \vec{p}_a = \frac{{^a\mathbf{A}_b}\vec{p}_b} {{\vec{a}^T_{3}}\vec{p}_b} + \frac{H}{Z_a}\frac{T_{_Z}}{D_{\pi, b}}\frac{{^a\mathbf{A}_b}\vec{p}_b} {{\vec{a}^T_{3}}\vec{p}_b} - \frac{H}{Z_a D_{\pi, b}}\vec{T}
\end{equation}
where $({{^a\mathbf{A}_b}\vec{p}_b}) / ({{\vec{a}^T_{3}}\vec{p}_b})$ is transformed pixel position of $\vec{p}_b$ on the first image due to the homography matrix, which is denoted as $\vec{p}_w$, and with definition of projective structure $\lambda = {H}/{Z_a}$ and epipole $\vec{e}_a = \vec{T}/{T_{_Z}}$ in the first image, the Eq~\ref{eq:13} can be simplified as:
\begin{equation}
\label{eq:14}
    \vec{p}_a = \vec{p}_w + \lambda \frac{T_{_Z}}{D_{\pi, b}}(\vec{p}_w - \vec{e}_a)
\end{equation}
where $\gamma_a$ is the projective structure of point $P$ in coordinate frame $a$ and can be expressed as $\gamma_a = {H}/{Z}$, and $\vec{e}_a$ is the epipole in the reference frame $a$ and defined as ${\vec{t}_a}/{T_{_Z}}$. With Eq~\ref{eq:14}, the displacement of pixel point $\vec{p}_a$ and $\vec{p}_b$ can be expressed as follows:
    \begin{equation}
    \begin{split}
	    \label{eq:3}
	    \Vec{p}_b - \Vec{p}_a& = (\Vec{p}_b - \Vec{p}_w) -\gamma_a\frac{T_{_Z}}{D_{\pi,b}}(\vec{p}_w - \vec{e}_a)\\
	    &=\vec{u}_{\pi} +  \vec{\mu}
	\end{split}
	\end{equation}
	where $\vec{u}_{\pi} =\Vec{p}_b - \Vec{p}_w $ defines the image displacement via the homography based on the plane $\pi$, and $\vec{\mu} = -\gamma_a{T_{_Z}}/{D_{\pi,b}}(\vec{p}_w - \vec{e}_a)$ gives the residual parallax displacement. Hence, it is called as \textit{Plane + Parallax}. 
	
	From the given Eq~\ref{eq:3} of residual parallax displacement, we can tell that the residual parallax vectors of rigid scene points intersect at the epipole $e$. Denote $\vec{p}_1$ and $\vec{p}_2$ are two points of the static scene on the first image. With elinimation of epipole in their residual parallax vectors, we will obtain the following equation:
	\begin{equation}
	\label{eq:4}
	    \vec{\mu}_1\gamma_2-\vec{\mu}_2\gamma_1 = \gamma_1\gamma_2\frac{T_{_Z}}{d_{\pi,b}}(\vec{p}_{w2} - \vec{p}_{w1})
	\end{equation}
	The Eq~\ref{eq:4} implies that the vectors on both sides are parallel, so that the \textit{relative structure} for a pair of points $\vec{p}_1$ and $\vec{p}_2$ can be computed as follows:
	\begin{equation}
	\label{eq:5}
	    \frac{\gamma_2}{\gamma_1} = \frac{\vec{\mu}_2^T(\Delta\vec{p}_w)_{\perp}}{\vec{\mu}_1^T(\Delta\vec{p}_w)_{\perp}}
	\end{equation}
	where $\Delta\vec{p}_w = \vec{p}_{w2} - \vec{p}_{w1}$ and $\vec{p}_{\perp}$ is an orthogonal vector to $\vec{p}$. Assume the projective structure $\gamma_1$ is known, then the structure for all other points in the image can be derived by the Eq~\ref{eq:5} without computing the epipole and camera movement.





\bibliography{Bibtex}

\begin{thebibliography}{10}
\expandafter\ifx\csname url\endcsname\relax
  \def\url#1{\texttt{#1}}\fi
\expandafter\ifx\csname urlprefix\endcsname\relax\def\urlprefix{URL }\fi
\expandafter\ifx\csname href\endcsname\relax
  \def\href#1#2{#2} \def\path#1{#1}\fi

\bibitem{LI2021108116}
S.~Li, J.~Shi, W.~Song, A.~Hao, H.~Qin,
  \href{https://www.sciencedirect.com/science/article/pii/S0031320321003034}{Hierarchical
  object relationship constrained monocular depth estimation.}, Pattern
  Recognition 120 (2021) 108116.
\newblock \href
  {http://dx.doi.org/https://doi.org/10.1016/j.patcog.2021.108116}
  {\path{doi:https://doi.org/10.1016/j.patcog.2021.108116}}.
\newline\urlprefix\url{https://www.sciencedirect.com/science/article/pii/S0031320321003034}

\bibitem{eigen2014depth}
D.~Eigen, C.~Puhrsch, R.~Fergus, Depth map prediction from a single image using
  a multi-scale deep network, in: Advances in Neural Information Processing
  Systems, 2014, pp. 2366--2374.

\bibitem{liu2015learning}
F.~Liu, C.~Shen, G.~Lin, I.~Reid, Learning depth from single monocular images
  using deep convolutional neural fields, IEEE Transactions on Pattern Analysis
  and Machine Intelligence 38~(10) (2015) 2024--2039.

\bibitem{luo2019every}
C.~Luo, Z.~Yang, P.~Wang, Y.~Wang, W.~Xu, R.~Nevatia, A.~Yuille, Every pixel
  counts ++: Joint learning of geometry and motion with 3d holistic
  understanding, IEEE Transactions on Pattern Analysis and Machine Intelligence
  42~(10) (2020) 2624--2641.
\newblock \href {http://dx.doi.org/10.1109/TPAMI.2019.2930258}
  {\path{doi:10.1109/TPAMI.2019.2930258}}.

\bibitem{godard2019digging}
C.~Godard, O.~Mac~Aodha, M.~Firman, G.~J. Brostow, Digging into self-supervised
  monocular depth estimation, in: Proceedings of the IEEE International
  Conference on Computer Vision (ICCV), 2019, pp. 3828--3838.

\bibitem{zhang2020depth}
K.~Zhang, J.~Xie, N.~Snavely, Q.~Chen, Depth sensing beyond lidar range, in:
  Proceedings of the IEEE/CVF Conference on Computer Vision and Pattern
  Recognition (CVPR), 2020, pp. 1692--1700.

\bibitem{irani1996parallax}
M.~Irani, P.~Anandan, Parallax geometry of pairs of points for 3d scene
  analysis, in: European Conference on Computer Vision, Springer, 1996, pp.
  17--30.

\bibitem{wulff2017optical}
J.~Wulff, L.~Sevilla-Lara, M.~J. Black, Optical flow in mostly rigid scenes,
  in: Proceedings of the IEEE Conference on Computer Vision and Pattern
  Recognition (CVPR), 2017, pp. 4671--4680.

\bibitem{sawhney19943d}
H.~S. Sawhney, 3d geometry from planar parallax, in: Proceedings of the
  IEEE/CVF Conference on Computer Vision and Pattern Recognition (CVPR),
  Vol.~94, 1994, pp. 929--934.

\bibitem{chaney2019learning}
K.~Chaney, A.~Zihao~Zhu, K.~Daniilidis, Learning event-based height from plane
  and parallax, in: Proceedings of the IEEE Conference on Computer Vision and
  Pattern Recognition Workshops, 2019, pp. 0--0.

\bibitem{geiger2012we}
[dataset], A.~Geiger, P.~Lenz, R.~Urtasun,
  \href{http://www.cvlibs.net/datasets/kitti}{Are we ready for autonomous
  driving? the kitti vision benchmark suite}, in: 2012 IEEE Conference on
  Computer Vision and Pattern Recognition (CVPR), 2012, pp. 3354--3361.
\newline\urlprefix\url{http://www.cvlibs.net/datasets/kitti}

\bibitem{Cordts2016Cityscapes}
[dataset], M.~Cordts, M.~Omran, S.~Ramos, T.~Rehfeld, M.~Enzweiler,
  R.~Benenson, U.~Franke, S.~Roth, B.~Schiele,
  \href{https://www.cityscapes-dataset.com}{The cityscapes dataset for semantic
  urban scene understanding}, in: Proceeding of the IEEE Conference on Computer
  Vision and Pattern Recognition (CVPR), 2016.
\newline\urlprefix\url{https://www.cityscapes-dataset.com}

\bibitem{eigen2015predicting}
D.~Eigen, R.~Fergus, Predicting depth, surface normals and semantic labels with
  a common multi-scale convolutional architecture, in: Proceedings of the IEEE
  International Conference on Computer Vision (ICCV), 2015, pp. 2650--2658.

\bibitem{GeoNet}
X.~Qi, R.~Liao, Z.~Liu, R.~Urtasun, J.~Jia, Geonet: Geometric neural network
  for joint depth and surface normal estimation, in: Proceedings of the IEEE
  Conference on Computer Vision and Pattern Recognition (CVPR), 2018, pp.
  283--291.

\bibitem{fu2018deep}
H.~Fu, M.~Gong, C.~Wang, K.~Batmanghelich, D.~Tao, Deep ordinal regression
  network for monocular depth estimation, in: Proceedings of the IEEE
  Conference on Computer Vision and Pattern Recognition (CVPR), 2018, pp.
  2002--2011.

\bibitem{XUE2021107901}
F.~Xue, J.~Cao, Y.~Zhou, F.~Sheng, Y.~Wang, A.~Ming,
  \href{https://www.sciencedirect.com/science/article/pii/S0031320321000881}{Boundary-induced
  and scene-aggregated network for monocular depth prediction}, Pattern
  Recognition 115 (2021) 107901.
\newblock \href
  {http://dx.doi.org/https://doi.org/10.1016/j.patcog.2021.107901}
  {\path{doi:https://doi.org/10.1016/j.patcog.2021.107901}}.
\newline\urlprefix\url{https://www.sciencedirect.com/science/article/pii/S0031320321000881}

\bibitem{ranftl2021vision}
R.~Ranftl, A.~Bochkovskiy, V.~Koltun, Vision transformers for dense prediction,
  in: Proceedings of the IEEE/CVF International Conference on Computer Vision
  (ICCV), 2021, pp. 12179--12188.

\bibitem{bhat2021adabins}
S.~F. Bhat, I.~Alhashim, P.~Wonka, Adabins: Depth estimation using adaptive
  bins, in: Proceedings of the IEEE/CVF Conference on Computer Vision and
  Pattern Recognition (CVPR), 2021, pp. 4009--4018.

\bibitem{YE2021107578}
X.~Ye, S.~Chen, R.~Xu,
  \href{https://www.sciencedirect.com/science/article/pii/S0031320320303812}{Dpnet:
  Detail-preserving network for high quality monocular depth estimation},
  Pattern Recognition 109 (2021) 107578.
\newblock \href
  {http://dx.doi.org/https://doi.org/10.1016/j.patcog.2020.107578}
  {\path{doi:https://doi.org/10.1016/j.patcog.2020.107578}}.
\newline\urlprefix\url{https://www.sciencedirect.com/science/article/pii/S0031320320303812}

\bibitem{garg2016unsupervised}
R.~Garg, B.~V. Kumar, G.~Carneiro, I.~Reid, Unsupervised cnn for single view
  depth estimation: Geometry to the rescue, in: European Conference on Computer
  Vision (ECCV), Springer, 2016, pp. 740--756.

\bibitem{godard2017unsupervised}
C.~Godard, O.~Mac~Aodha, G.~J. Brostow, Unsupervised monocular depth estimation
  with left-right consistency, in: Proceedings of the IEEE Conference on
  Computer Vision and Pattern Recognition (CVPR), 2017, pp. 270--279.

\bibitem{zhou2017unsupervised}
T.~Zhou, M.~Brown, N.~Snavely, D.~G. Lowe, Unsupervised learning of depth and
  ego-motion from video, in: Proceedings of the IEEE Conference on Computer
  Vision and Pattern Recognition (CVPR), 2017, pp. 1851--1858.

\bibitem{wang2018learning}
C.~Wang, J.~Miguel~Buenaposada, R.~Zhu, S.~Lucey, Learning depth from monocular
  videos using direct methods, in: Proceedings of the IEEE Conference on
  Computer Vision and Pattern Recognition (CVPR), 2018, pp. 2022--2030.

\bibitem{watson2021temporal}
J.~Watson, O.~Mac~Aodha, V.~Prisacariu, G.~Brostow, M.~Firman, The temporal
  opportunist: Self-supervised multi-frame monocular depth, in: Proceedings of
  the IEEE/CVF Conference on Computer Vision and Pattern Recognition (CVPR),
  2021, pp. 1164--1174.

\bibitem{casser2019depth}
V.~Casser, S.~Pirk, R.~Mahjourian, A.~Angelova, Depth prediction without the
  sensors: Leveraging structure for unsupervised learning from monocular
  videos, in: Proceedings of the AAAI Conference on Artificial Intelligence,
  Vol.~33, 2019, pp. 8001--8008.

\bibitem{yang2021learning}
G.~Yang, D.~Ramanan, Learning to segment rigid motions from two frames, in:
  Proceedings of the IEEE/CVF Conference on Computer Vision and Pattern
  Recognition (CVPR), 2021, pp. 1266--1275.

\bibitem{zhu2019improving}
Y.~Zhu, K.~Sapra, F.~A. Reda, K.~J. Shih, S.~Newsam, A.~Tao, B.~Catanzaro,
  Improving semantic segmentation via video propagation and label relaxation,
  in: Proceedings of the IEEE Conference on Computer Vision and Pattern
  Recognition (CVPR), 2019, pp. 8856--8865.

\bibitem{li2019learning}
Z.~Li, T.~Dekel, F.~Cole, R.~Tucker, N.~Snavely, C.~Liu, W.~T. Freeman,
  Learning the depths of moving people by watching frozen people, in:
  Proceedings of the IEEE Conference on Computer Vision and Pattern Recognition
  (CVPR), 2019, pp. 4521--4530.

\bibitem{zhang2020resnest}
H.~Zhang, C.~Wu, Z.~Zhang, Y.~Zhu, H.~Lin, Z.~Zhang, Y.~Sun, T.~He, J.~Mueller,
  R.~Manmatha, et~al., Resnest: Split-attention networks, arXiv preprint
  arXiv:2004.08955.

\bibitem{uhrig2017sparsity}
J.~Uhrig, N.~Schneider, L.~Schneider, U.~Franke, T.~Brox, A.~Geiger, Sparsity
  invariant cnns, in: 2017 International Conference on 3D Vision (3DV), IEEE,
  2017, pp. 11--20.

\bibitem{watson2019self}
J.~Watson, M.~Firman, G.~J. Brostow, D.~Turmukhambetov, Self-supervised
  monocular depth hints, in: Proceedings of the IEEE International Conference
  on Computer Vision (ICCV), 2019, pp. 2162--2171.

\bibitem{he2016deep}
K.~He, X.~Zhang, S.~Ren, J.~Sun, Deep residual learning for image recognition,
  in: Proceedings of the IEEE Conference on Computer Vision and Pattern
  Recognition (CVPR), 2016, pp. 770--778.

\bibitem{xie2017aggregated}
S.~Xie, R.~Girshick, P.~Doll{\'a}r, Z.~Tu, K.~He, Aggregated residual
  transformations for deep neural networks, in: Proceedings of the IEEE
  Conference on Computer Vision and Pattern Recognition (CVPR), 2017, pp.
  1492--1500.

\bibitem{yin2018geonet}
Z.~Yin, J.~Shi, Geonet: Unsupervised learning of dense depth, optical flow and
  camera pose, in: Proceedings of the IEEE Conference on Computer Vision and
  Pattern Recognition (CVPR), 2018, pp. 1983--1992.

\bibitem{ranjan2019competitive}
A.~Ranjan, V.~Jampani, L.~Balles, K.~Kim, D.~Sun, J.~Wulff, M.~J. Black,
  Competitive collaboration: Joint unsupervised learning of depth, camera
  motion, optical flow and motion segmentation, in: Proceedings of the IEEE
  conference on Computer Vision and Pattern Recognition (CVPR), 2019, pp.
  12240--12249.

\bibitem{pilzer2018unsupervised}
A.~Pilzer, D.~Xu, M.~Puscas, E.~Ricci, N.~Sebe, Unsupervised adversarial depth
  estimation using cycled generative networks, in: 2018 International
  Conference on 3D Vision (3DV), IEEE, 2018, pp. 587--595.

\bibitem{laina2016deeper}
I.~Laina, C.~Rupprecht, V.~Belagiannis, F.~Tombari, N.~Navab, Deeper depth
  prediction with fully convolutional residual networks, in: 2016 Fourth
  international conference on 3D vision (3DV), IEEE, 2016, pp. 239--248.

\bibitem{xu2018pad}
D.~Xu, W.~Ouyang, X.~Wang, N.~Sebe, Pad-net: Multi-tasks guided
  prediction-and-distillation network for simultaneous depth estimation and
  scene parsing, in: Proceedings of the IEEE Conference on Computer Vision and
  Pattern Recognition, 2018, pp. 675--684.

\bibitem{zhang2018joint}
Z.~Zhang, Z.~Cui, C.~Xu, Z.~Jie, X.~Li, J.~Yang, Joint task-recursive learning
  for semantic segmentation and depth estimation, in: Proceedings of the
  European Conference on Computer Vision (ECCV), 2018, pp. 235--251.

\bibitem{wang2020sdc}
L.~Wang, J.~Zhang, O.~Wang, Z.~Lin, H.~Lu, Sdc-depth: Semantic
  divide-and-conquer network for monocular depth estimation, in: Proceedings of
  the IEEE/CVF Conference on Computer Vision and Pattern Recognition, 2020, pp.
  541--550.

\end{thebibliography}


  \begin{wrapfigure}{l}{0.16\textwidth} 
	\includegraphics[width=0.13\textwidth, clip,keepaspectratio]{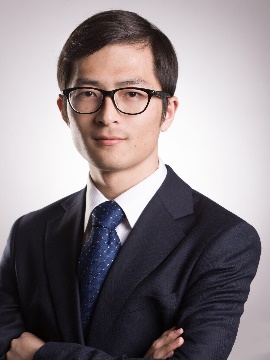}
\end{wrapfigure}
\par\textbf{Hao Xing} received his B.S. degree in mechanical engineering from Hefei University of Technology, Anhui, China (2014) and M.S. degree at Technical University of Munich, Munich, Germany (2019). He is currently pursuing a Ph.D. in computer science department at Technical University of Munich. Research interests: Computer Vision and Robotics. \par \break

\begin{wrapfigure}{l}{0.175\textwidth} 
	\includegraphics[width=0.13\textwidth, clip,keepaspectratio]{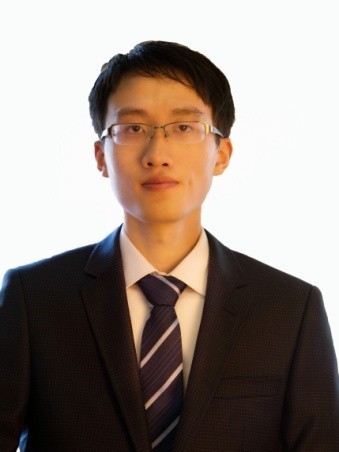}
\end{wrapfigure}\par\textbf{YIFAN CAO} received the B.S. degree in mechanical engineering (mechatronics) from Duisburg-Essen university in 2017. He is currently focusing on master’s degree in mechanical engineering at the Technical University of Munich. His research direction includes computer vision and deep learning.\par  
\vskip 1cm

\begin{wrapfigure}{l}{0.16\textwidth} 
	\includegraphics[width=0.15\textwidth, clip,keepaspectratio]{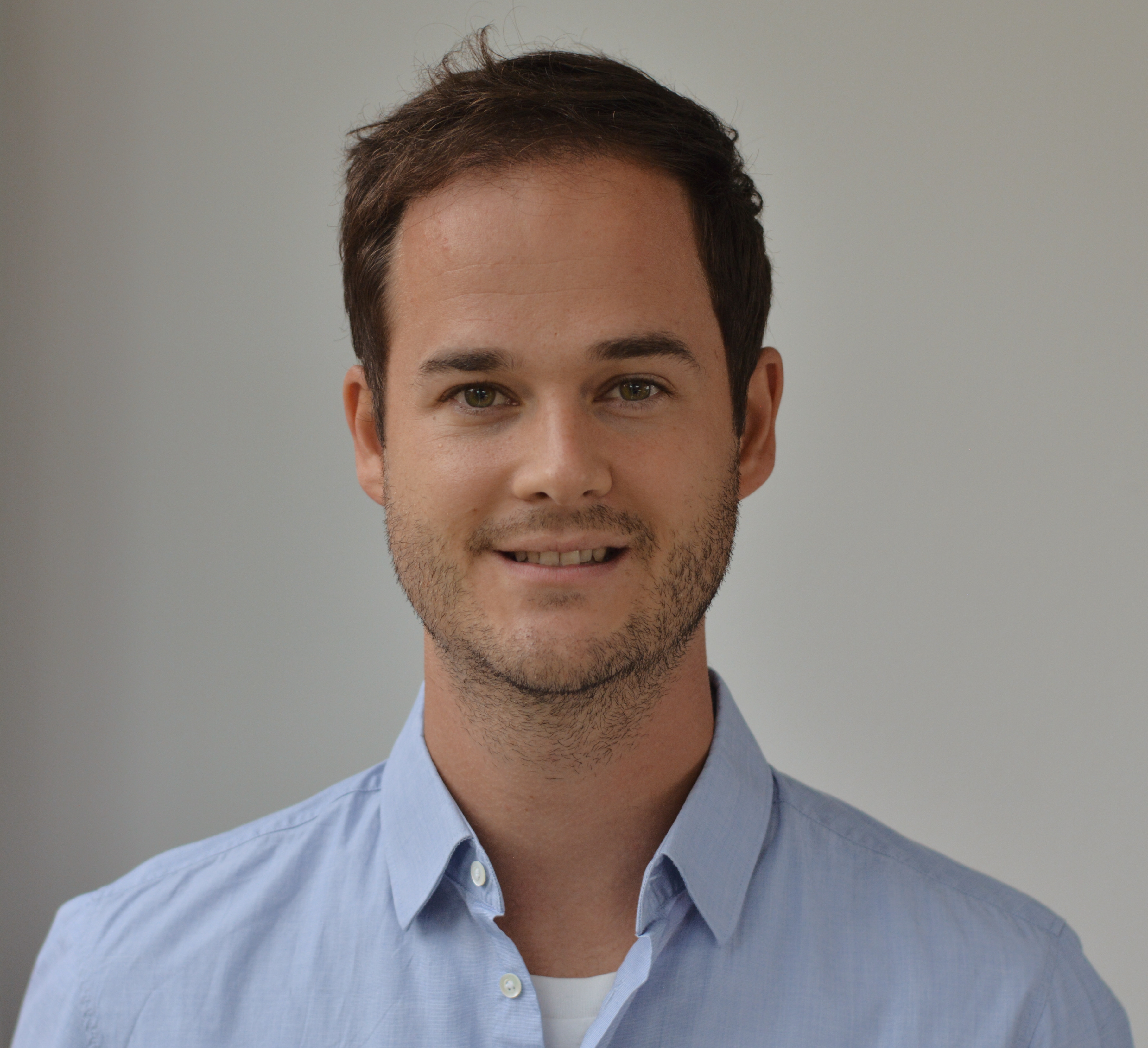}
\end{wrapfigure}\par\textbf{Maximilian Biber} is a Machine Learning Engineer with a master’s degree in Robotics, Cognition, Intelligence from the Technical University Munich (TUM). In addition to that he has a background in entrepreneurship and innovation from Unternehmer TUM and currently works in the AI consultancy industry.\par  
\vskip 1cm\

\begin{wrapfigure}{l}{0.16\textwidth} 
	\includegraphics[width=0.13\textwidth, clip,keepaspectratio]{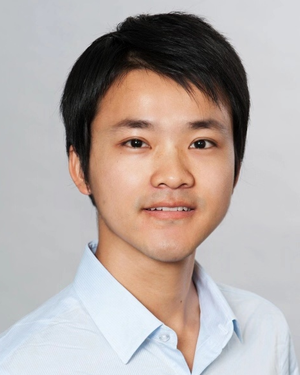}
\end{wrapfigure}
\par\textbf{MINGCHUAN ZHOU} received a Ph.D. degree in computer science from the Technical University of Munich, Munich, Germany, in 2020. He was a postdoc at the Helmholtz Centre Munich from 2019 to 2021. He is currently an assistant professor leading the multi-scale robotic manipulation lab for agriculture in Zhejiang University. \par
\vskip 1cm

\begin{wrapfigure}{l}{0.15\textwidth} 
	\includegraphics[width=0.13\textwidth, clip,keepaspectratio]{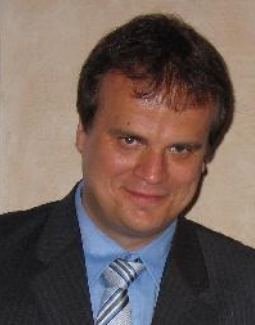}
\end{wrapfigure}
\par\textbf{DARIUS BURSCHKA} received a Ph.D. degree in department of electrical engineering at Technical University of Munich, Munich, Germany, in 1998 and acquired his postdoctoral in Center for Computational Vision and Control at Yale University, USA, in 1999. He is currently an associate professor in department of computer science at Technical University of Munich. \par

\end{document}